\documentclass[twoside, 11pt]{article}

\usepackage{jmlr2e}
\usepackage[algo2e,ruled]{algorithm2e}
\usepackage{graphicx} 
\usepackage{subfigure}
\usepackage{hyperref,multirow,float}
\usepackage{xspace}
\usepackage{amsmath}
\usepackage{steinmetz}

\usepackage{rotating} 

\jmlrheading{1}{2014}{1-xx}{11/20}{xx/14}{Dhruv Mahajan, S. Sathiya Keerthi and S. Sundararajan}

\ShortHeadings{Distributed block coordinate descent for $l_1$ regularized classifiers}{Mahajan, Keerthi and Sundararajan}
\firstpageno{1}

\newcommand*{\Scale}[2][4]{\scalebox{#1}{\ensuremath{#2}}}%
\def\ccomp{C_{\mbox{comp}}^P}
\def\ccomm{C_{\mbox{comm}}^P}

\def\dbcd{{\textsc{DBCD}}}
\def\dbcds{{\textsc{DBCD-S}}}
\def\fpa{{\textsc{FPA}}}
\def\grock{{\textsc{GROCK}}}
\def\pcdns{{\textsc{PCD-S}}}
\def\pcdnr{{\textsc{PCD-R}}}
\def\pcdn{{\textsc{PCD}}}
\def\dbcdr{{\textsc{DBCD-R}}}
\def\richtarik{{\textsc{HYDRA}}}
\def\admm{{\textsc{ADMM}}}

\begin{document}

\title{A distributed block coordinate descent method for training $\boldmath{\Scale[1]{l_1}}$ regularized linear classifiers}

\author{\name Dhruv Mahajan \email dhrumaha@microsoft.com \\
       \addr Cloud \& Information Services Lab\\
       Microsoft\\
       Mountain View, CA 94043, USA
       \AND
       \name S. Sathiya Keerthi \email keerthi@microsoft.com \\
       \addr Cloud \& Information Services Lab\\
       Microsoft\\
       Mountain View, CA 94043, USA
       \AND
       \name S. Sundararajan \email ssrajan@microsoft.com \\
       \addr Microsoft Research \\
       Bangalore, India}

\editor{xxx}

\maketitle
\begin{abstract}
Distributed training of $l_1$ regularized classifiers has received great attention recently. Most existing methods approach this problem by taking steps obtained from approximating the objective by a quadratic approximation that is decoupled at the individual variable level. These methods are designed for multicore and MPI platforms where communication costs are low. They are inefficient on systems such as Hadoop running on a cluster of commodity machines where communication costs are substantial. In this paper we design a distributed algorithm for $l_1$ regularization that is much better suited for such systems than existing algorithms. A careful cost analysis is used to support these points and motivate our method. The main idea of our algorithm is to do block optimization of many variables on the actual objective function within each computing node; this increases the computational cost per step that is matched with the communication cost, and decreases the number of outer iterations, thus yielding a faster overall method. Distributed Gauss-Seidel and Gauss-Southwell greedy schemes are used for choosing variables to update in each step. We establish global convergence theory for our algorithm, including Q-linear rate of convergence. Experiments on two benchmark problems show our method to be much faster than existing methods.
\end{abstract}

\begin{keywords}
  Distributed learning, $l_1$ regularization
\end{keywords}

\def\wsp{w_{S_p^t}^t}
\def\wspp{w_{S_p^t}^{t+1}}
\def\wbp{w_{B_p}}
\def\wbphat{{\hat w}_{B_p}}
\def\wbpt{w_{B_p}^t}
\def\wbptp{w_{B_p}^{t+1}}
\def\dsp{d_{S_p^t}^t}
\def\dbp{d_{B_p}^t}
\def\Xsp{X_{S_p^t}}
\def\Xbp{X_{B_p}}
\def\gsp{g_{S_p^t}}
\def\gbp{g_{B_p}}
\def\gbpt{g_{B_p}^t}
\def\bpbar{\bar{B}_p}
\def\wbpbar{w_{\bpbar}^t}
\def\wbarpt{\bar{w}_{B_p}^t}
\def\wbarpts{\bar{w}_{S_p^t}^t}

\def\grad{\nabla}
\def\Cone{{\cal{C}}^1}
\def\Ctwo{{\cal{C}}^2}
\def\wbar{\bar{w}}
\def\gbar{\bar{g}}
\def\Lhat{\hat{L}}
\def\fhat{\hat{f}}
\def\Fhat{\hat{F}}
\def\ghat{\hat{g}}
\def\what{\hat{w}}
\def\Hhat{\hat{H}}
\def\xihat{\hat{\xi}}
\def\delhat{\hat{\delta}}
\def\gamhat{\hat{\gamma}}
\def\jbar{\bar{j}}
\def\Abar{{\bar{A}}}
\def\Hbar{{\bar{H}}}
\def\winf{w_{\infty}}
\def\N{\cal{M}}
\def\kaprime{\kappa^\prime}
\def\xbar{\bar{x}}

\def\xstar{x^\star}

\def\wtilde{\tilde{w}}
\def\kappap{\kappa^\prime}
\def\what{\hat{w}}
\def\dhat{\hat{d}}
\def\mysgn{\operatorname{sgn}}
\def\ftilde{\tilde{f}}
\def\khat{\hat{k}}
\def\defs{\stackrel{\text{def}}{=}}

\def\ttilde{\tilde{t}}
\def\that{\hat{t}}

\section{Introduction}
\label{sec:intro}

The design of sparse linear classifiers using $l_1$ regularization is an important problem that has received great attention in recent years. This is due to its value in scenarios where the number of features is large and the classifier representation needs to be kept compact. Big data is becoming common nowadays. For example, in online advertising one comes across datasets with about a billion examples and a billion features. A substantial fraction of the features is usually irrelevant; and, $l_1$ regularization offers a systematic way to choose the small fraction of relevant features and form the classifier using them. In the future, one can foresee even bigger sized datasets to arise in this and other applications. For such big data, distributed storage of data over a cluster of commodity machines becomes necessary.  Thus, fast training of $l_1$ regularized classifiers over distributed data is an important problem.

A number of algorithms have been recently proposed for parallel and distributed training of $l_1$ regularized classifiers; see Section~\ref{sec:rw} for a review. Most of these algorithms are based on coordinate-descent and they assume the data to be feature-partitioned. They are designed for multicore and MPI platforms in which data communication costs are negligible. These platforms are usually equipped with only a small number of computing nodes. Distributed systems, e.g., Hadoop running on a cluster of commodity machines, are better for employing a large number of nodes and hence, for inexpensive handling of big data. However, in such systems, communication costs are high; current methods for $l_1$ regularization are not optimally designed for such systems. {\em In this paper we develop a distributed block coordinate descent (DBCD) method that is efficient on distributed platforms in which communication costs are high.}

The paper is organized as follows. Most methods (including the current ones and the one we propose) fit into a generic algorithm format that we describe in Section~\ref{sec:generic}. This gives a clear view of existing methods and allows us to motivate the new method. In Section~\ref{sec:rw} we discuss the key related work in some detail. The analysis of computation and communication costs in Section~\ref{sec:motiv} motivates our DBCD method. In Section~\ref{sec:dbcd} we describe the DBCD method in detail and prove its convergence. Experiments comparing our method with several existing methods on a few large scale datasets are given in Section~\ref{sec:expts}. These experiments strongly demonstrate the efficiency of one version of our method that chooses update variables greedily. This best version of the DBCD method is described in Section~\ref{sec:recom}. Section~\ref{sec:conc} contains some concluding comments.

\section{A generic algorithm}
\label{sec:generic}

The generic algorithm format allows us to explain the roles of key elements of various methods and point out how new choices for the steps can lead to a better design. Before describing it, we first formulate the $l_1$ regularization problem.

{\bf Problem formulation.}
Let $w$ be the weight vector with $m$ variables, $w_j$, $j=1,\ldots,m$, and $x_i\in R^m$ denote the $i$-th example. 
Note that we have denoted vector components by subscripts, e.g., $w_j$ is the $j$-th component of $w$; we have also used subscripts for indexing examples, e.g., $x_i$ is the $i$-th example, which itself is a vector. But this will not cause confusion anywhere.
A linear classifier produces the output vector $y_i=w^Tx_i$. The loss is a nonlinear convex function applied on the outputs. For binary class label $c_i\in\{1,-1\}$, the loss is given by $\ell(y_i;c_i)$. Let us simply view $\ell(y_i;c_i)$ as a function of $y_i$ with $c_i$ acting as a parameter. We will assume that $\ell$ is non-negative and convex, $\ell\in\Cone$, the class of continuously differentiable functions, and that $\ell^\prime$ is Lipschitz continuous\footnote{{\small A function $h$ is Lipschitz continuous if there exists a (Lipschitz) constant $L\ge 0$ such that $\|h(a)-h(b)\|\le L \|a-b\| \;\; \forall \; a,b$.}}. Loss functions such as least squares loss, logistic loss, SVM squared hinge loss and Huber loss satisfy these assumptions. The total loss function, $f:R^m\to R$ is $f(w)=\frac{1}{n}\sum_i \ell(y_i;c_i)$.
Let $u$ be the $l_1$ regularizer given by $u(w) = \lambda \sum_j |w_j|$, where $\lambda>0$ is the regularization constant. Our aim is to solve the problem
\begin{equation}
\min_{w\in R^m} F(w) = f(w) + u(w).
\label{minF}
\end{equation}
Let $g=\grad f$. The optimality conditions for (\ref{minF}) are:
\begin{equation}
\forall j: \;\; g_j+\lambda\; \mbox{sign} (w_j) = 0 \; \mbox{if}\;\; |w_j|>0; \;\; |g_j|\le\lambda \; \mbox{if}\;\; w_j=0.
\label{viol}
\end{equation}

Let there be $n$ training examples and let $X$ denote the $n\times m$ data matrix, whose $i$-th row is $x_i^T$. For problems with a large number of features, it is natural to randomly partition the columns of $X$ and place the parts in $P$ computing nodes. Let  $\{B_p\}_{p=1}^P$ denote this partition of ${\N} = \{1,\ldots,m\}$, i.e., $B_p \subset {\N} \;\forall p$ and $\cup_p B_p = \N$. 
{\em We will assume that this feature partitioning is given and that all algorithms operate within that constraint.} The variables associated with a particular partition get placed in one node. Given a subset of variables $S$, let $X_S$ be the submatrix of $X$ containing the columns corresponding to $S$. For a vector $z\in R^m$, $z_S$ will denote the vector containing the components of $z$ corresponding to $S$.

{\bf Generic algorithm.}
Algorithm 1 gives the generic algorithm.
Items such as $B_p$, $S_p^t$, $\wbp$, $\dbp$, $\Xbp$ stay local in node $p$ and do not need to be communicated. Step (d) can be carried out using an {\it AllReduce} operation~\citep{alekh2013} over the nodes and then $y$ becomes available in all the nodes. The gradient sub-vector $\gbpt$ (which is needed for solving~(\ref{Fapprox})) can then be computed locally as $\gbpt = \Xbp^T b$ where $b\in R^n$ is a vector with $\{\ell^\prime(y_i)\}$ as its components.

\begin{algorithm2e}
\caption{A generic distributed algorithm\label{GA}}
Choose $w^0$ and compute $y^0=Xw^0$\;
\For{$t=0,1 \ldots$}{
    \For{$p=1,\ldots, P$}{
          (a) Select a subset of variables, $S_p^t\subset B_p$\;
          (b) Form $f_p^t(\wbp)$, an approximation of $f$ and solve (exactly or approximately):
          \begin{equation}
          \min f_p^t(\wbp) + u(\wbp) \;\;\; \mbox{s.t.} \;\;\; w_j = w^t_j \; \forall \; j\not\in B_p\setminus{S_p^t}
          \label{Fapprox}
          \end{equation}
          to get $\wbarpt$ and set direction: $\dbp=\wbarpt-\wbpt$\;
          (c) Choose $\alpha^t$ and update: $\wbptp = \wbpt + \alpha^t\dbp$\;
     }
     (d) Update $y^{t+1} = y^t + \alpha^t \sum_p \Xbp\dbp$\;
     (e) Terminate if optimality conditions hold\;
}
\end{algorithm2e}


{\bf Step (a) - variable sampling.} Some choices are: 
\begin{itemize}
\item 
{\bf (a.1)} random selection~\citep{bradley2011, richtarik2012}; 
\item
{\bf (a.2)} random cyclic: over a set of consecutive iterations $(t)$ all variables are touched once~\citep{bian2013}; 
\item 
{\bf (a.3)}  greedy: always choose a set of variables that, in some sense violate (\ref{viol}) the most at the current iterate~\citep{peng2013, facchinei2013}; and, 
\item 
{\bf (a.4)} greedy selection using the Gauss-Southwell rule~\citep{tseng2009, yun2011}.
\end{itemize}

{\bf Step (b) - function approximation.} Most methods choose a quadratic approximation that is decoupled at the individual variable level:
\begin{equation}
f_p^t(\wbpt) = \sum_{j\in B_p} g_{j}(w^t) (w_j-w^t_j) + \frac{L_j}{2} (w_j-w^t_j)^2
\label{quad1}
\end{equation}
The main advantages of (\ref{quad1}) are its simplicity and closed-form minimization when used in~(\ref{Fapprox}). Choices for $L^j$ that have been tried are: 
\begin{itemize}
\item 
{\bf (b.1)} $L_j=$ a Lipschitz constant for $g_j$~\citep{bradley2011,peng2013}; 
\item 
{\bf (b.2)} $L_j=$ a large enough bound on the Lipschitz constant for $g_j$ to suit the sampling in step (a)~\citep{richtarik2012}; 
\item 
{\bf (b.3)} adaptive adjustment of $L_j$~\citep{facchinei2013}; and 
\item 
{\bf (b.4)} $L_j=H_{jj}^t$, the $j$-th diagonal term of the Hessian at $w^t$~\citep{bian2013}.
\end{itemize}

{\bf Step (c) - step size.} The choices are: 
\begin{itemize}
\item 
{\bf (c.1)} always fix $\alpha^t=1$~\citep{bradley2011,richtarik2012,peng2013}; 
\item 
{\bf (c.2)} use stochastic approximation ideas to choose $\{\alpha^t\}$ so that $\sum_t(\alpha^t)^2<\infty$ and $\sum_t|\alpha^t|=\infty$~\citep{facchinei2013}; and 
\item 
{\bf (c.3)} choose $\alpha^t$ by line search that is directly tied to the optimization of $F$ in (\ref{minF})~\citep{bian2013}.
\end{itemize}

To understand the role of the various choices better, let us focus on the use of (\ref{quad1}) for $f_p^t$. Algorithm 1 may not converge to the optimal solution due to one of the following decisions: (i) choosing too many variables ($|S_p^t|$ large) for parallel updating in step (a); (ii) choosing small values for the proximal coefficient $L_j$ in step (b); and (iii) not controlling $\alpha^t$ to be sufficiently small in step (c). This is because each of the above has the potential to cause large step sizes leading to increases in $F$ value and, if this happens uncontrolled at all iterations then convergence to the minimum cannot occur. Different methods control against these by making suitable choices in the steps.

The choice made for step (c) gives a nice delineation of methods. With {\bf (c.1)}, one has to do a suitable mix of large enough $L_j$ and small enough $|S_p^t|$. Choice {\bf (c.2)} is better since the proper control of $\{\alpha^t\}\rightarrow 0$ takes care of convergence; however, for good practical performance, $L_j$ and $\alpha^t$ need to be carefully adapted, which is usually messy. Choice {\bf (c.3)} is good in many ways: it leads to monotone decrease in $F$; it is good theoretically and practically; and, it allows both, small $L_j$ as well as large $|S_p^t|$ without hindering convergence. Except for~\citet{bian2013}, \citet{tseng2009} and \citet{yun2011}\footnote{Among these three works, \citet{tseng2009} and \citet{yun2011} mainly focus on general theory and little on distributed implementation.}, {\bf (c.3)} has been unused in other methods because it is considered as `not-in-line' with a proper parallel approach as it requires a separate $\alpha^t$ determination step requiring distributed computations and also needing $F$ computations for several $\alpha^t$ values within one $t$. With line search, the actual implementation of Algorithm 1 merges steps {\bf (c)} and {\bf (d)} and so it deviates slightly from the flow of Algorithm 1. Specifically, we compute $\delta y = \sum_p \Xbp\dbp$ before line search using AllReduce. Then each node can compute $f$ at any $\alpha$ locally using $y+\alpha\, \delta y$. Only a scalar corresponding to the $l_1$ regularization term needs to be communicated for each $\alpha$. This means that the communication cost associated with line search is minimal.\footnote{Later, in Section~\ref{sec:motiv} when we write costs, we write it to be consistent with Algorithm 1. The total cost of all the steps is the same for the implementation described here for line search. For genericity sake, we keep Algorithm 1 as it is even for the line search case. The actual details of the implementation for the line search case will become clear when we layout the final algorithm in Section~\ref{sec:recom}.} But truly, the slightly increased computation and communication costs is amply made up by a reduction in the number of iterations to reach sufficient optimality. So we go with the choice {\bf (c.3)} in our method. 

The choice of (\ref{quad1}) for $f_p^t$ in step (b) is pretty much unanimously used in all previous works. While this is fine for communication friendly platforms such as multicore and MPI, it is not the right choice when communication costs are high. Such a setting permits more per-node computation time, and there is much to be gained by using a more complex $f_p^t$. We propose the use of a function $f_p^t$ that couples the variables in $S_p^t$. We also advocate an approximate solution of (\ref{Fapprox}) (e.g., a few rounds of coordinate descent within each node) in order to control the computation time.

Crucial gains are also possible via resorting to the greedy choices, {\bf (a.3)} and {\bf (a.4)} for choosing $S_p^t$. On the other hand, with methods based on {\bf (c.1)}, one has to be careful in using {\bf (a.3)}: apart from difficulties in establishing convergence, practical performance can also be bad, as we show in Section~\ref{sec:expts}.




{\bf Contributions.}
Following are our main contributions.
\begin{enumerate}
\item We provide a cost analysis that brings out the computation and communication costs of Algorithm 1 clearly for different methods. In the process we motivate the need for new efficient methods suited to communication heavy settings.
\item We make careful choices for the three steps of Algorithm 1, leading to the development of a distributed block coordinate descent (DBCD) method that is very efficient on distributed platforms with high communication cost.
\item We establish convergence theory for our method using the results of~\citet{tseng2009} and \citet{yun2011}. It is worth noting the following: (a) though \citet{tseng2009} and \citet{yun2011} cover algorithms using quadratic approximations for the total loss, we use a simple trick to apply them to general nonlinear approximations, thus bringing more power to their results; and (b) even these two works use only (\ref{quad1}) in their implementations whereas we work with more powerful approximations that couple features.
\item We give an experimental evaluation that shows the strong performance of DBCD against key current methods in scenarios where communication cost is significant. Based on the experiments we make a final recommendation for the best method to employ for such scenarios.
\end{enumerate}

\section{Related Work}
\label{sec:rw}


Our interest is mainly in parallel/distributed computing methods. There are many parallel algorithms targeting a single machine having multi-cores with shared memory~\citep{bradley2011, richtarik2013, bian2013, peng2013}. In contrast, there exist only a few efficient algorithms to solve (\ref{minF}) when the data is distributed \citep{richtarik2013a, Ravazzi2013} and communication is an important aspect to consider. In this setting, the problem (\ref{minF}) can be solved in several ways depending on how the data is distributed across machines~\citep{peng2013,boyd2011}: (A) example (horizontal) split, (B) feature (vertical) split and (C) combined example and feature split (a block of examples/features per node). While methods such as distributed \textsc{FISTA}~\citep{peng2013} or \textsc{ADMM}~\citep{boyd2011} are useful for (A), the block splitting method~\citep{parikh2013} is useful for (C). We are interested in (B), and the most relevant and important class of methods is parallel/distributed coordinate descent methods, as abstracted in Algorithm 1. Most of these methods set $f_p^t$ in step (b) of Algorithm 1 to be a quadratic approximation that is decoupled at the individual variable level. Table~\ref{tab:methodscomp} compares these methods along various dimensions.\footnote{Although our method will be presented only in Section 5, we include our method's properties in the last row of Table~\ref{tab:methodscomp}. This helps to easily compare our method against the rest.} 

Most dimensions arise naturally from the steps of Algorithm~1, as explained in Section 2. Two important points to note are: (i) except~\citet{richtarik2013a} and our method, none of these methods target and sufficiently discuss distributed setting involving communication and, (ii) from a practical view point, it is difficult to ensure stability and get good speed-up with no line search and non-monotone methods. For example, methods such as~\citet{bradley2011,richtarik2012,richtarik2013,peng2013} that do not do line search are shown to have the monotone property only in expectation and that too only under certain conditions. Furthermore, variable selection rules, proximal coefficients and other method-specific parameter settings play important roles in achieving monotone convergence and improved efficiency. As we show in Section 6, our method and the parallel coordinate descent Newton method~\citep{bian2013} (see below for a discussion) enjoy robustness to various settings and come out as clear winners. 

It is beyond the scope of this paper to give a more detailed discussion, beyond Table~\ref{tab:methodscomp}, of the methods from a theoretical convergence perspective on various assumptions and conditions under which results hold. We only briefly describe and comment on them below. \vspace*{0.05in}

\noindent{\bf Generic Coordinate Descent Method~\citep{scherrer2012a,scherrer2012}} \citet{scherrer2012a} and \citet{scherrer2012} presented an abstract framework for coordinate descent methods (\textsc{GenCD}) suitable for parallel computing environments. Several coordinate descent algorithms such as stochastic coordinate descent~\citep{shwartz2011}, \textsc{Shotgun}~\citep{bradley2011} and \textsc{GROCK}~\citep{peng2013} are covered by \textsc{GenCD}. \textsc{GROCK} is a {thread greedy} algorithm~\citep{scherrer2012a} in which the variables are selected greedily using gradient information. One important issue is that algorithms such as \textsc{Shotgun} and \textsc{GROCK} may not converge in practice due to their non-monotone nature with no line search; we faced convergence issues on some datasets in our experiments with \textsc{GROCK} (see Section 6). Therefore, the practical utility of such algorithms is limited without ensuring necessary descent property through certain spectral radius conditions on the data matrix. \vspace*{0.05in}

\noindent{\bf Distributed Coordinate Descent Method~\citep{richtarik2013a}} 
The multi-core parallel coordinate descent method of \citet{richtarik2012} is a much refined version of \textsc{GenCD} with careful choices for steps (a)-(c) of Algorithm 1 and a supporting stochastic convergence theory. \citet{richtarik2013a} extended this to the distributed setting; so, this method is more relevant to this paper. With no line search, their algorithm \textsc{HYDRA} (Hybrid coordinate descent) has (expected) descent property only for certain sampling types of selecting variables and $L_j$ values. One key issue is setting the right $L_j$ values for good performance. Doing this accurately is a costly operation; on the other hand, inaccurate setting using cheaper computations (e.g., using the number of non-zero elements as suggested in their work) results in slower convergence (see Section 6). \vspace*{0.05in}

\citet{neco2013} suggest another variant of parallel coordinate descent in which all the variables are updated in each iteration. \textsc{HYDRA} and \textsc{GROCK} can be considered as two key, distinct methods that represent the set of methods discussed above. So, in our analysis as well as experimental comparisons in the rest of the paper, we do not consider the methods in this set other than these two.

\noindent{\bf Flexible Parallel Algorithm (\textsc{FPA})~\citep{facchinei2013} } This method has some similarities with our method in terms of the approximate function optimized at the nodes. Though~\citet{facchinei2013} suggest several approximations, they use only (\ref{quad1}) in its final implementation. More importantly, \textsc{FPA} is a non-monotone method using a stochastic approximation step size rule. Tuning this step size rule along with the proximal parameter $L_j$ to ensure convergence and speed-up is hard. (In Section 6 we conduct experiments to show this.) Unlike our method, \textsc{FPA}'s inner optimization stopping criterion is unverifiable (for e.g., with (\ref{ours})); also, \textsc{FPA} does not address the communication cost issue. \vspace*{0.05in}


\noindent{\bf Parallel Coordinate Descent Newton (\textsc{PCD})~\citep{bian2013}}
One key difference between other methods discussed above and our \textsc{DBCD} method is the use of line search. Note that the \textsc{PCD} method can be seen as a special case of \textsc{DBCD} (see Section 5.1). In DBCD, we optimize per-node block variables jointly, and perform line search across the blocks of variables; as shown later in our experimental results, this has the advantage of reducing the number of outer iterations, and overall wall clock time due to reduced communication time (compared to \textsc{PCD}). \vspace*{0.05in}

\noindent{\bf Synchronized Parallel Algorithm~\citep{patriksson1998fp}} \citet{patriksson1998fp} proposed a Jacobi type synchronous parallel algorithm with line search using a generic cost approximation (\textsc{CA}) framework for differentiable objective functions~\citep{patriksson1998}. Its local linear rate of convergence results hold only for a class of strong monotone \textsc{CA} functions. If we view the approximation function, $f_p^t$ as a mapping that is dependent on $w^t$, \citet{patriksson1998fp} requires this mapping to be continuous, which is unnecessarily restrictive. 


\noindent{\bf \textsc{ADMM} Methods} Alternating direction method of multipliers is a generic and popular distributed computing method. It does not fit into the format of Algorithm~1. This method can be used to solve (\ref{minF}) in different data splitting scenarios~\citep{boyd2011,parikh2013}. Several variants of global convergence and rate of convergence (e.g., $O(\frac{1}{k})$) results exist under different weak/strong convexity assumptions on the two terms of the objective function~\citep{deng2012,deng2013}. Recently, an accelerated version of \textsc{ADMM}~\citep{goldstein2013} derived using the ideas of Nesterov's accelerated gradient method~\citep{nesterov2012} has been proposed; this method has dual objective function convergence rate of $O(\frac{1}{k^2})$ under a strong convexity assumption. \textsc{ADMM} performance is quite good when the augmented Lagrangian parameter is set to the right value; however, getting a reasonably good value comes with computational cost. In Section 6 we evaluate our method and find it to be much faster. \vspace*{0.05in}

Based on the above study of related work, we choose \textsc{HYDRA}, \textsc{GROCK}, \textsc{PCD} and \textsc{FPA} as the main methods for analysis and comparison with our method.\footnote{In the experiments of Section~\ref{sec:expts}, we also include ADMM.} Thus, Table~1 gives various dimensions only for these methods.

\begin{sidewaystable}
\centering

\caption{Properties of selected methods that fit into the format of Algorithm~1. Methods: \richtarik~\citep{richtarik2013a}, \grock~~\citep{peng2013}, \fpa~~\citep{facchinei2013}, \pcdn~~\citep{bian2013}. }
\label{tab:methodscomp}
{\small
\begin{tabular}{|c|c|c|c|c|c|c|c|} \hline
{\bf Method} & {\bf Is ${\mathbf F(w^t)}$} & {\bf Are limits}                       & {\bf How is ${\mathbf S_p^t}$} & {\bf Basis for}                & {\bf How is}                      & {\bf Convergence} & {\bf Convergence} \\
             & {\bf monotone?}             & {\bf forced on ${\mathbf |S_p^t|}$?}   & {\bf chosen?}                  & {\bf choosing ${\mathbf L_j}$} & {\bf ${\mathbf \alpha^t}$ chosen} & {\bf type}        & {\bf rate} \\
\hline
\multicolumn{8}{c}{{\bf Existing methods}} \\ \hline 
\richtarik & No & No, if $L_j$ is & Random & Lipschitz bound for $g_j$ & Fixed & Stochastic & Linear \\
           &    & varied suitably &        & suited to $S_p^t$ choice  &      &                &        \\ \hline
\grock & No & Yes     & Greedy & Lipschitz bound for $g_j$ & Fixed & Deterministic & Sub-linear \\ \hline
\fpa & No & No & Random & Lipschitz bound for $g_j$ & Adaptive & Deterministic & None \\ \hline
\pcdn & Yes & No & Random & Hessian~diagonal & Armijo      & Stochastic     & Sub-linear \\ 
      &     &    &        &                  & line search &                &            \\ \hline
\multicolumn{8}{c}{{\bf Our method}} \\ \hline 
\textsc{DBCD} & Yes & No & Random/Greedy & Free & Armijo      & Deterministic  & Locally linear \\
              &     &    &               &      & line search &                &                \\ \hline
\end{tabular}
}

\end{sidewaystable}

\section{DBCD method: Motivation}
\label{sec:motiv}

Our goal is to develop an efficient distributed learning method that jointly optimizes the costs involved in the various steps of the algorithm. We observed that the methods discussed in Section~\ref{sec:rw} lack this careful optimization in one or more steps, resulting in inferior performance. In this section, we present a detailed cost analysis study and motivate our optimization strategy that forms the basis for our DBCD method.  

{\bf Remark.} A non-expert reader could find this section hard to read in the first reading because of two reasons: (a) it requires a decent understanding of several methods covered in Section~\ref{sec:rw}; and (b) it requires knowledge of the details of our method. For this sake, let us give the main ideas of this section in a nutshell. (i) The cost of Algorithm 1 can be written as $T^P(\ccomp + \ccomm)$ where $P$ denotes the number of nodes, $T^P$ is the number of outer iterations\footnote{For practical purposes, one can view $T^P$ as the number of outer iterations needed to reach a specified closeness to the optimal objective function value. We will say this more precisely in Section~\ref{sec:expts}.}, and, $\ccomp$ and $\ccomm$ respectively denote the computation and communication costs per-iteration. (ii) In communication heavy situations, existing algorithms have $\ccomp \ll \ccomm$. (iii) Our method aims to improve overall efficiency by making each iteration more complex ($\ccomp$ is increased) and, in the process, making $T^P$ much smaller. {\it A serious reader can return to study the details behind these ideas after reading Section~\ref{sec:dbcd}.}

Following Section~\ref{sec:rw}, we 
select the following methods for our study:
(1) \richtarik~\citep{richtarik2013a}, (2) \grock~(Greedy coordinate-block)~\citep{peng2013}, (3) \fpa~(Flexible Parallel Algorithm)~\citep{facchinei2013}, and (4) \pcdn~(Parallel Coordinate Descent Newton method)~\citep{bian2013}.
We will use the above mentioned abbreviations for the methods in the rest of the paper. 


Let $nz$ and $|S| = \sum_p |S^t_p|$ denote the number of non-zero entries in the data matrix $X$ and the number of variables updated in each iteration respectively. To keep the analysis simple, we make the homogeneity assumption that the number of non-zero data elements in each node is $nz/P$. Let $\beta (\gg 1)$ be the relative computation to communication speed in the given distributed system; more precisely, it is the ratio of the times associated with communicating a floating point number and performing one floating point operation.  Recall that $n$, $m$ and $P$ denote the number of examples, features and nodes respectively. Table~\ref{tab:stepcost} gives cost expressions for different steps of the algorithm in one outer iteration. Here $c_1$, $c_2$, $c_3$, $c_4$ and $c_5$ are method dependent parameters.  
Table~\ref{tab:methodcost} gives the cost constants for various methods.\footnote{As in Table~\ref{tab:methodscomp}, for ease of comparison, we also list the cost constants for our method (three variations), in Table~\ref{tab:methodcost}. The details for them will become clear in Section~\ref{sec:dbcd}.}
We briefly discuss different costs below.

\begin{table}[h]
\caption{Cost of various steps of Algorithm~\ref{GA}. $\ccomp$ and $\ccomm$ are respectively, the sums of costs in the computation and communication rows.}
\label{tab:stepcost}
\begin{center}
\begin{tabular}{|c|c|c|c|c|c|}
\hline
Cost & \multicolumn{4}{|c|}{Steps of Algorithm~\ref{GA}} \\
\hline
& Step a & Step b & Step c & Step d \\
& {\small Variable sampling} & {\small Inner optimization} & {\small Choosing step size} & {\small Updating output} \\
\hline
Computation & $c_1\frac{nz}{P}$  & $c_2\frac{nz}{P}\frac{|S|}{m}$  &  $c_3|S| + c_4n$  & $c_5\frac{nz}{P}\frac{|S|}{m}$ \\
\hline
Communication\protect\footnotemark & -  & - &  $c_4\beta logP$  & $\beta n logP$ \\
\hline
\end{tabular}
\end{center}
\end{table}

\footnotetext{Note that the communication latency cost (time for the first byte to reach the destination) is ignored in the communication cost expressions because it is dominated by the throughput cost for large $n$. Moreover, our {\it{AllReduce}} is a non-pipelined version of the implementation in~\citet{alekh2013}.}

\begin{table}[h]
\caption{Cost parameter values and costs for different methods. $q$ lies in the range: $1 \leq q \leq \frac{m}{|S|}$. $R$ and $S$ refer to variable selection schemes for step (a); see Section~\ref{sec:dbcd}. \pcdn~uses the $R$ scheme and so it can also be referred to as \pcdnr. Typically $\tau_{ls}$, the number of $\alpha$ values tried in line search, is very small; in our experiments we found that on average it is not more than $10$. Therefore all methods have pretty much the same communication cost per iteration.}
\vspace*{-0.1in}
\label{tab:methodcost}
\begin{center}
\begin{tabular}{|c|c|c|c|c|c|c|c|}
\hline
Method & $c_1$ & $c_2$ & $c_3$ & $c_4$ & $c_5$ & Computation & Communication \\
       &       &       &       &       &       & cost per iteration & cost per iteration \\
\hline
\multicolumn{8}{c}{{\bf Existing methods}} \\
\hline
\richtarik & 0 & 1 & 1 & 0 & 1 & $2\frac{nz}{P}\frac{|S|}{m} + |S|$   & $\beta n \log P$\\
\hline
\grock & 1 & $q$ & 1 & 0 & $q$ & $\frac{nz}{P} + 2q\frac{nz}{P}\frac{|S|}{m} + |S|$ & $\beta n \log P$ \\
\hline
\fpa & 1 & $q$ & 1 & 1 & $q$  & $\frac{nz}{P} + 2q\frac{nz}{P}\frac{|S|}{m} + |S| + n$ & $\beta (n+1) \log P$\\
\hline
\pcdn & 0 & 1 & $\tau_{ls}$ & $\tau_{ls}$ & 1 & $2\frac{nz}{P}\frac{|S|}{m} + \tau_{ls}|S| + \tau_{ls}n$ & $\beta (n+\tau_{ls}) \log P$ \\
\hline
\multicolumn{8}{c}{{\bf Variations of our method}} \\
\hline
\pcdns & 1 & $q$ & $\tau_{ls}$ & $\tau_{ls}$ & $q$ & $\frac{nz}{P} + 2q\frac{nz}{P}\frac{|S|}{m} + \tau_{ls}|S| + \tau_{ls}n$ & $\beta (n+\tau_{ls}) \log P$ \\
\hline
\dbcdr & 0 & $k$ & $\tau_{ls}$ & $\tau_{ls}$ & 1 & $(k+1)\frac{nz}{P}\frac{|S|}{m} + \tau_{ls}|S| + \tau_{ls}n$ & $\beta (n+\tau_{ls}) \log P$ \\
\hline
\dbcds & 1 & $kq$ & $\tau_{ls}$ & $\tau_{ls}$ & $q$ & $\frac{nz}{P} + q(k+1)\frac{nz}{P}\frac{|S|}{m} + \tau_{ls}|S| + \tau_{ls}n$ & $\beta (n+\tau_{ls}) \log P$ \\
\hline
\end{tabular}
\end{center}
\end{table}

\noindent{\bf{Step a:}} Methods like our \dbcds\footnote{The DBCD and PCD methods have two variants, R and S corresponding to different ways of implementing step a; these will be discussed in Section~\ref{sec:dbcd}.}, \grock, \fpa~ and \pcdns~ need to calculate the gradient and model update to determine which variables to update. Hence, they need to go through the whole data once ($c_1=1$). On the other hand \richtarik,~\pcdn~ and \dbcdr~ select
variables randomly or in a cyclic order. As a result variable subset selection cost is negligible for them ($c_1=0$).\vspace{0.05in}\\
{\bf{Step b:}} All the methods except \dbcds~ and \dbcdr~ use the decoupled quadratic approximation~(\ref{quad1}). For \dbcdr~ and \dbcds,~ an additional factor of $k$ comes in $c_2$ since we do $k$ inner cycles of \textsc{CDN} in each iteration. \richtarik, \pcdn~ and \dbcdr~ do a random or cyclic selection of variables. Hence, a factor of $\frac{|S|}{m}$ comes in the cost since only a subset $|S|$ of variables is updated in each iteration. However, methods that do selection of variables based on the magnitude of update or expected objective function decrease (\dbcds, \grock, \fpa~ and \pcdns) favour variables with low sparsity. As a result, $c_2$ for these methods has an additional factor $q$ where $1 \leq q \leq \frac{m}{|S|}$.\vspace{0.05in}\\
{\bf{Step c:}} For methods that do not use line-search, $c_3 = 1$ and $c_4 = 0$\footnote{{\small For \fpa, $c_4=1$ since objective function needs to be computed to automatically set the proximal term parameter.}}. The overall cost is $|S|$ to update the variables. For methods like \dbcds, \dbcdr, \pcdn~ and \pcdns~ that do line-search, $c_3 = c_4 = \tau_{ls}$ where $\tau_{ls}$ is the average number of steps ($\alpha$ values tried) in one line search. For each line search step, we need to recompute the loss function which involves going over $n$ examples once. Moreover, \textit{AllReduce} step needs to be performed to sum over the distributed $l_1$ regularizer term. Hence, an  additional $\beta log P$ cost is incurred to communicate the local regularizer. As pointed out in~\citet{bian2013}, $\tau_{ls}$ can increase with $P$; but it is still negligible compared to $n$.\vspace{0.05in}\\
{\bf{Step d:}} This step involves computing and doing \textit{AllReduce }on updated local predictions to get the global prediction vector for the next iteration and is common for all the methods.

The analysis given above is only for $\ccomp$ and $\ccomm$, the computation and communication costs in one iteration. If $T^P$ is the number of iterations to reach a certain optimality tolerance, then the total cost of Algorithm 1 is: $C^P = T^P(\ccomp + \ccomm)$. For $P$ nodes, speed-up is given by $C^1/C^P$. To illustrate the ill-effects of communication cost, let us take the method of~\citet{richtarik2013}. For illustration, take the case of $|S|=P$, i.e., one variable is updated per node per iteration. For large $P$, $C^P\approx T^P\ccomm = T^P\; \beta n \log P$; both $\beta$ and $n$ are large in the distributed setting. On the other hand, for $P=1$, $\ccomm=0$ and $C^P=\ccomp \approx \frac{nz}{m} $. Thus
${\mbox speed up} \;\; = \;\; \frac{T^1}{T^P} \frac{C^1}{C^P} \;\; = \;\; \frac{T^1}{T^P} \frac{\frac{nz}{m}}{\beta n \log P}$.
\citet{richtarik2013} show that $T^1/T^P$ increases nicely with $P$. But, the term $\beta n$ in the denominator of $C^1/C^P$ has a severe detrimental effect. Unless a special distributed system with efficient communication is used, speed up has to necessarily suffer. When the training data is huge and so the data is forced to reside in distributed nodes, {\it the right question to ask is not whether we get great speed up, but to ask which method is the fastest}. Given this, we ask how various choices in the steps of Algorithm 1 can be made to decrease $C^P$. Suppose we devise choices such that (a) $\ccomp$ is increased while still remaining in the zone where $\ccomp \ll \ccomm$, and (b) in the process, $T^P$ is decreased greatly, then $C^P$ can be decreased. The basic idea of our method is to use a more complex $f_p^t$ than the simple quadratic in (\ref{quad1}), due to which, $T^P$ becomes much smaller. The use of line search, {\bf (c.3)} for step {\bf c} aids this further. We see in Table~\ref{tab:methodcost} that, \dbcdr~ and \dbcds~ have the maximum computational cost. On the other hand, communication cost is more or less the same for all the methods (except for few scalars in the line search step) and dominates the cost. In Section~\ref{sec:expts}, we will see on various datasets how, by doing more computation, our methods reduce $T^P$ substantially over the other methods while incurring a small computation overhead (relative to communication) per iteration. These will become amply clear in Section~\ref{sec:expts}; see, for example, Table~\ref{tab:numiter} in that section.


\section{DBCD method}
\label{sec:dbcd}

The DBCD method that we propose fits into the general format of Algorithm 1. It is actually {\it a class of algorithms} that allows various possibilities for steps (a), (b) and (c). Below we lay out these possibilities and establish convergence theory for our method. We also show connection to other methods on aspects such as function approximation, variable selection, etc.

\subsection{Function approximation}
\label{subsec:func}

Let us begin with step (b). There are three key items involved: (i) what are some of the choices of approximate functions possible, used by our methods and others? (ii) what is the stopping criterion for the inner optimization (i.e., local problem), and, (iii) what is the method used to solve the inner optimization? We discuss all these details below. We stress the main point that, unlike previous methods, we allow $f_p^t$ to be non-quadratic and also to be a joint function of the variables in $\wbp$. We first describe a general set of properties that $f_p^t$ must satisfy, and then discuss specific instantiations that satisfy these properties.

{\bf P1}. $f_p^t\in\Cone$; $g_p^t=\grad f_p^t$ is Lipschitz continuous, with the Lipschitz constant uniformly bounded over all $t$; $f_p^t$ is strongly convex (uniformly in $t$), i.e., $\exists\; \mu>0$ such that $f_p^t-\frac{\mu}{2} \| \wbp \|^2$ is convex; and, $f_p^t$ is gradient consistent with $f$ at $\wbpt$, i.e., $g_p^t(\wbpt) = g_{B_p}(w^t)$.

This assumption is not restrictive. Gradient consistency is essential because it is the property that connects $f_p^t$ to $f$ and ensures that a solution of (\ref{Fapprox}) will make $\dbp$ a descent direction for $F$ at $\wbpt$, thus paving the way for a decrease in $F$ at step (c).
Strong convexity is a technical requirement that is needed for establishing sufficient decrease in $F$ in each step of Algorithm 1. Our experiments indicate that it is sufficient to set $\mu$ to be a very small positive value. Lipschitz continuity is another technical condition that is needed for ensuring boundedness of various quantities; also, it is easily satisfied by most loss functions. Let us now discuss some good ways of choosing $f_p^t$. {\em For all these instantiations, a proximal term is added to get the strong convexity required by} {\bf P1}.

{\bf Proximal-Jacobi.} We can follow the classical Jacobi method in choosing $f_p^t$ to be the restriction of $f$ to $\wsp$, with the remaining variables fixed at their values in $w^t$. Let $\bpbar$ denote the complement of $B_p$, i.e., the set of variables associated with nodes other than $p$. Thus we set
\begin{equation}
f_p^t(\wbp) = f(\wbp,\wbpbar) + \frac{\mu}{2} \| \wbp - \wbpt \|^2
\label{ours}
\end{equation}
where $\mu>0$ is the proximal constant. It is worth pointing out that, since each node $p$ keeps a copy of the full classifier output vector $y$ aggregated over all the nodes, the computation of $f_p^t$ and $g_p^t$ due to changes in $\wbp$ can be locally computed in node $p$. Thus the solution of (\ref{Fapprox}) is local to node $p$ and so step (b) of Algorithm 1 can be executed in parallel for all $p$.

{\bf Block GLMNET.} GLMNET~\citep{yuan2012, friedman2010} is a sequential coordinate descent method that has been demonstrated to be very promising for the sequential solution of $l_1$ regularized problems with logistic loss. At each iteration, GLMNET minimizes the second order Taylor series of $f$ at $w^t$, followed by line search along the direction generated by this minimizer. We can make a distributed version by choosing $f_p^t$ to be the second order Taylor series approximation of $f(\wbp,\wbpbar)$ restricted to $\wbp$ while keeping $w_{\bpbar}$ fixed at $\wbpbar$.

{\bf Block \textsc{L-BFGS}.} One can keep a limited history of $\wbpt$ and $\gbpt$ and use an $L-BFGS$ approach to build a second order approximation of $f$ in each iteration to form $f_p^t$.

{\bf Decoupled quadratic.} Like in existing methods we can also form a quadratic approximation of $f$ that decouples at the variable level. If the second order term is based on the diagonal elements of the Hessian at $w^t$, then the PCDN algorithm given in~\citet{bian2013} can be viewed as a special case of our DBCD method. PCDN~\citep{bian2013} is based on Gauss-Seidel variable selection. But it can also be used in combination with the distributed greedy scheme that we propose in Subsection 5.2 below.

{\bf Approximate stopping.}
In step (b) of Algorithm 1 we mentioned the possibility of approximately solving~(\ref{Fapprox}). This is irrelevant for previous methods which solve individual variable level quadratic optimization in closed form, but very relevant to our method. Here we propose an approximate relative stopping criterion and later, in Subsection 5.4, also give convergence theory to support it.

Let $\partial u_j$ be the set of sub-gradients of the regularizer term $u_j = \lambda |w_j|$, i.e.,
\begin{equation}
\partial u_j = [-\lambda,\lambda] \;\; \mbox{if} \; w_j=0; \;\; \lambda \; \mbox{sign} (w_j) \;\; \mbox{if} \; w_j\not=0.
\label{subg}
\end{equation}
A point $\wbarpt$ is optimal for (\ref{Fapprox}) if, at that point,
\begin{equation}
(g_p^t)_j + \xi_j = 0, \;\; \mbox{for some} \;\; \xi_j\in\partial u_j \;\; \forall \; j\in S_p^t.
\label{opt}
\end{equation}
An approximate stopping condition can be derived by choosing a tolerance $\epsilon>0$ and requiring that, for each $j\in S_p^t$ there exists $\xi_j\in\partial u_j$ such that
\begin{equation}
\delta^j = (g_p^t)_j + \xi_j, \;\; |\delta_j| \le \epsilon |d^t_j|  \;\; \forall \; j\in S_p^t
\label{appopt}
\end{equation}

{\bf Method used for solving~(\ref{Fapprox}).}
Now (\ref{Fapprox}) is an $l_1$ regularized problem restricted to $w_{S_p^t}$. It has to be solved within node $p$ using a suitable sequential method. Going by the state of the art for sequential solution of such problems~\citep{yuan2010} we use the coordinate-descent method described in~\citet{yuan2010} for solving~(\ref{Fapprox}).

\subsection{Variable selection}
\label{subsec:varsel}

Let us now turn to step (a) of Algorithm 1. We propose two schemes for variable selection, i.e., choosing $S_p^t\subset B_p$.

{\bf Gauss-Seidel scheme.}
In this scheme, we form cycles - each cycle consists of a set of consecutive iterations - while making sure that every variable is touched once in each cycle. We implement a cycle as follows. Let $\tau$ denote the iteration where a cycle starts. Choose a positive integer $T$ ($T$ may change with each cycle). For each $p$, randomly partition $B_p$ into $T$ equal parts: $\{S_p^t\}_{t=\tau}^{\tau+T-1}$. Use these variable selections to do $T$ iterations. {\em Henceforth, we refer to this scheme as the $R$-scheme.}

{\bf Distributed greedy scheme.}
This is a greedy scheme which is purely distributed and so more specific than the Gauss-Southwell schemes in~\citet{tseng2009}.\footnote{{\small Yet, our distributed greedy scheme can be shown to imply the Gauss-Southwell-$q$ rule for a certain parameter setting. See the appendix for details.}} In each iteration, our scheme chooses variables based on how badly~(\ref{viol}) is violated for various $j$. For one $j$, an expression of this violation is as follows. Let $g^t$ and $H^t$ denote, respectively, the gradient and Hessian at $w^t$. Form the following one variable quadratic approximation:
\begin{eqnarray}
q_j(w_j) = g^t_j (w_j - w^t_j) + \frac{1}{2} (H_{jj}^t+\nu) (w_j - w^t_j)^2 + \nonumber \\
\lambda |w_j| - \lambda|w^t_j|
\label{quad}
\end{eqnarray}
where $\nu$ is a small positive constant. Let ${\bar q}_j$ denote the optimal objective function value obtained by minimizing $q_j(w_j)$ over all $w_j$. Since $q_j(w^t_j)=0$, clearly ${\bar q}_j\le 0$. The more negative ${\bar q}_j$  is, the better it is to choose $j$.

Our distributed greedy scheme first chooses a working set size WSS and then, in each node $p$, it chooses the top WSS variables from $B_p$ according to smallness of ${\bar q}_j$, to form $S_p^t$. {\em Hereafter, we refer to this scheme as the $S$-scheme.}

It is worth pointing out that, our distributed greedy scheme requires more computation than the Gauss-Seidel scheme. However, since the increased computation is local, non-heavy and communication is the real bottleneck, it is not a worrisome factor.


\subsection{Line search}
\label{subsec:ls}

Line search (step (c) of Algorithm 1) forms an important component for making good decrease in $F$ at each iteration. For non-differentiable optimization, there are several ways of doing line search. For our context, \citet{tseng2009} and~\citet{patriksson1998} give two good ways of doing line search based on Armijo backtracking rule. In this paper we use ideas from the former. Let $\beta$ and $\sigma$ be real parameters in the interval $(0,1)$. (We use the standard choices, $\beta=0.5$ and $\sigma=0.01$.) We choose $\alpha^t$ to be the largest element of $\{\beta^k\}_{k=0,1,\ldots}$ satisfying
\begin{eqnarray}
F(w^t + \alpha^t d^t) \le F(w^t) + \alpha^t \sigma \Delta^t, \label{ls1} \\
\Delta^t \defs (g^t)^T d^t + \lambda u(w^t+d^t) - \lambda u(w^t). \label{ls2}
\end{eqnarray}

\subsection{Convergence}
\label{subsec:conv}

We now establish convergence for the class of algorithmic choices discussed in Subections 5.1-5.3. To do this, we make use of the results of~\citet{tseng2009}. An interesting aspect of this use is that, while the results of~\citet{tseng2009} are stated only for $f_p^t$ being quadratic, we employ a simple trick that lets us apply the results to our algorithm which involves non-quadratic approximations.

Apart from the conditions in {\bf P1} (see Subection 5.1) we need one other technical assumption.

{\bf P2.} For any given $t$, $\wbp$ and $\wbphat$, $\exists$ a positive definite matrix $\Hhat\ge\mu I$ (note: $\Hhat$ can depend on $t$, $\wbp$ and $\wbphat$) such that
\begin{equation}
g_p^t(\wbp) - g_p^t(\wbphat) = \Hhat (\wbp - \wbphat)
\label{mvt}
\end{equation}

Except {\it Proximal-Jacobi}, the other instantiations of $f_p^t$ mentioned in Subection 5.1 are quadratic functions; for these, $g_p^t$ is a linear function and so (\ref{mvt}) holds trivially. Let us turn to {\it Proximal-Jacobi}. If $f_p^t\in\Ctwo$, the class of twice continuously differentiable functions, then {\bf P2} follows directly from mean value theorem; note that, since $f_p^t-\frac{\mu}{2} \|w\|^2$ is convex, $H_p\ge \mu I$ at any point, where $H_p$ is the Hessian of $f_p^t$. Thus {\bf P2} easily holds for least squares loss and logistic loss.
Now consider the SVM squared hinge loss, $\ell(y_i;c_i) = 0.5(\max \{0, 1-y_ic_i\})^2$, which is not in $\Ctwo$. {\bf P2} holds for it because $g=\sum_i \ell^\prime (y_i;c_i) x_i$ and, for any two real numbers $z_1, z_2$, $\ell^\prime(z_1;c_i)-\ell^\prime(z_2;c_i) = \kappa (z_1,z_2,c_i) (z_1-z_2)$ where $0 \le \kappa(z_1,z_2,c_i) \le 1$.

The main convergence theorem can now be stated. Its proof is given in the appendix.

{\bf Theorem 1.} Suppose, in Algorithm 1: (i) step (a) is done via the Gauss-Seidel or distributed greedy schemes of Subection 5.2; (ii) $f_p^t$ in step (b) satisfies {\bf P1} and {\bf P2}; (iii) (\ref{appopt}) is used to terminate (\ref{Fapprox}) with $\epsilon=\mu/2$ (where $\mu$ is as in {\bf P1}); and (iv) in step (c), $\alpha^t$ is chosen via Armijo backtracking of Subection 5.3. Then Algorithm 1 is well defined and produces a sequence, $\{w^t\}$ such that any accumulation point of $\{w^t\}$ is a solution of (\ref{minF}). If, in addition, the total loss, $f$ is strongly convex, then $\{F(w^t)\}$ converges Q-linearly and $\{w^t\}$ converges at least R-linearly.\footnote{{\small See chapter 9 of~\citet{ortega1970} for definitions of Q-linear and R-linear convergence.}}

%

\section{Experimental Evaluation}
\label{sec:expts}

In this section, we present experimental results on real-world datasets. We compare our methods with several state of the art methods, in particular, those analyzed in Section~\ref{sec:motiv} (see the methods in the first column of Table 3) together with \admm, the accelerated alternating direction method of multipliers~\citep{goldstein2013}. 
To the best of our knowledge, such a detailed study has not been done for parallel and distributed $l_1$ regularized solutions in terms of (a) accuracy and solution optimality performance, (b) variable selection schemes, (c) computation versus communication time and (d) solution sparsity. The results demonstrate the effectiveness of our methods in terms of total (computation + communication) time on both accuracy and objective function measures.

\subsection{Experimental Setup}
\label{subsec:setup}

\noindent{\bf Datasets:} We conducted our experiments on two popular benchmark datasets \textsc{KDD} and \textsc{URL}\footnote{{\small See \url{http://www.csie.ntu.edu.tw/~cjlin/libsvmtools/datasets/}. We refer to kdd2010 (algebra) dataset as \textsc{KDD}.}}. \textsc{KDD} has $n=8.41\times 10^6$, $m=20.21\times 10^6$ and $nz=0.31\times 10^9$. \textsc{URL} has $n=2.00\times 10^6$, $m=3.23\times 10^6$ and $nz=0.22\times 10^9$.
These datasets have sufficiently interesting characteristics of having a large number of examples and features such that (1) feature partitioning, (2) $l_1$ regularization and (3) communication are important. \vspace*{0.05in}


\noindent{\bf Methods and Metrics:} We evaluate the performance of all the methods using (a) Area Under Precision-Recall Curve (AUPRC)~\citep{alekh2013} and (b) Relative Function Value Difference (RFVD) as a function of time taken. RFVD is computed as $\log(\frac{F(w^t)-F^{*}}{F^{*}})$ where $F^{*}$ is taken as the best value obtained across the methods after a long duration. We stopped each method after 800 outer iterations. We also report per node computation time statistics and sparsity pattern behavior of all the methods. \vspace*{0.05in}

\noindent{\bf Parameter Settings:} We experimented with the $\lambda$ values of $(1.23\times 10^{-5},1.37\times 10^{-6},4.6\times 10^{-7})$ and $(7.27\times 10^{-6},2.42\times 10^{-6},9\times 10^{-8})$ for the \textit{KDD} and \textit{URL} datasets respectively. These values are chosen in such a way that they are centered around the respective optimal $\lambda$ value and have good sparsity variations over the optimal solution. With respect to Algorithm 1, the working set size (WSS) per node and number of nodes ($P$) are common across all the methods. We set WSS in terms of the fraction ($r$) of the number of features per node, i.e., WSS=$r m/P$. Note that WSS will change with $P$ for a given fraction $r$. For \textit{KDD} and \textit{URL}, we used three $r$ values $(0.01,0.1,0.25)$ and $(0.001,0.01,0.1)$ respectively. We experimented with $P = 25, 100$.
Also, $r$ does not play a role in ADMM since all variables are optimized in each node. \vspace*{0.05in}

\noindent{\bf Platform:} We ran all our experiments on a Hadoop cluster with $379$ nodes and 10 Gbit interconnect speed. Each node has Intel (R) Xeon (R) E5-2450L (2 processors) running at 1.8 GHz and 192 GB RAM. (Though the datasets can fit in this memory configuration, our intention is to test the performance in a distributed setting.) All our implementations were done in $C\#$ including our binary tree \textit{AllReduce} support~\citep{alekh2013} on Hadoop.


\subsection{Method Specific Parameter Settings}
\label{subsec:par}

We discuss method specific parameter setting used in our experiments and associated practical implications.

Let us begin with \admm. We use the feature partitioning formulation of ADMM described in Subsection 8.3 of \citet{boyd2011}.
\admm~does not fit into the format of Algorithm 1, but the communication cost per outer iteration is comparable to the other methods that fit into Algorithm 1.
In \admm, the augmented Lagrangian parameter ($\rho$) plays an important role in getting good performance. In particular, the number of iterations required by ADMM for convergence is very sensitive with respect to $\rho$. While many schemes have been discussed in the literature~\citep{boyd2011} we found that selecting $\rho$ using the objective function value gave a good estimate; we selected $\rho^*$ from a handful of values with ADMM run for $10$ iterations (i.e., not full training) for each $\rho$ value tried.{\footnote{These initial ``tuning" iterations are not counted against the limit of 800 we set for the number of iterations. Thus, for ADMM, the total number of iterations can go higher than 800.} However, this step incurred some computational/communication time. In our time plots shown later, the late start of ADMM results is due to this cost. Note that this minimal number of iterations was required to get a decent $\rho^*$. \vspace*{0.05in}

\noindent{\bf Choice of ${\mathbf\mu}$ and $k$:} To get a practical implementation that gives good performance in our method, we deviate slightly from the conditions of Theorem 1. First, we find that the proximal term does not contribute usefully to the progress of the algorithm (see the left side plot in Figure 1). So we choose to set $\mu$ to a small value, e.g., $\mu=10^{-12}$. Second, we replace the stopping condition (\ref{appopt}) by simply using a fixed number of cycles of coordinate descent to minimize $f_p^t$. The right side plot in Figure 1 shows the effect of number of cycles, $k$. We found that a good choice for the number of cycles is $10$ and we used this value in all our experiments.

For \grock, \fpa~and \richtarik~we set the constants $(L_j)$ as suggested in the respective papers. Unfortunately, we found \grock~to be either unstable and diverging or extremely slow. The left side plot in Figure~\ref{fig:Issues} depicts these behaviors. The solid red line shows the divergence case. \fpa~requires an additional parameter ($\gamma$) setting for the stochastic approximation step size rule. Our experience is that setting right values for these parameters to get good performance can be tricky and highly dataset dependent. The right side plot in Figure~\ref{fig:Issues} shows the extremely slow convergence behavior of \fpa. Therefore, we do not include \grock~and \fpa~further in our study.

\begin{figure}[t]
\hspace{-0in}
\includegraphics[width=0.48\linewidth]{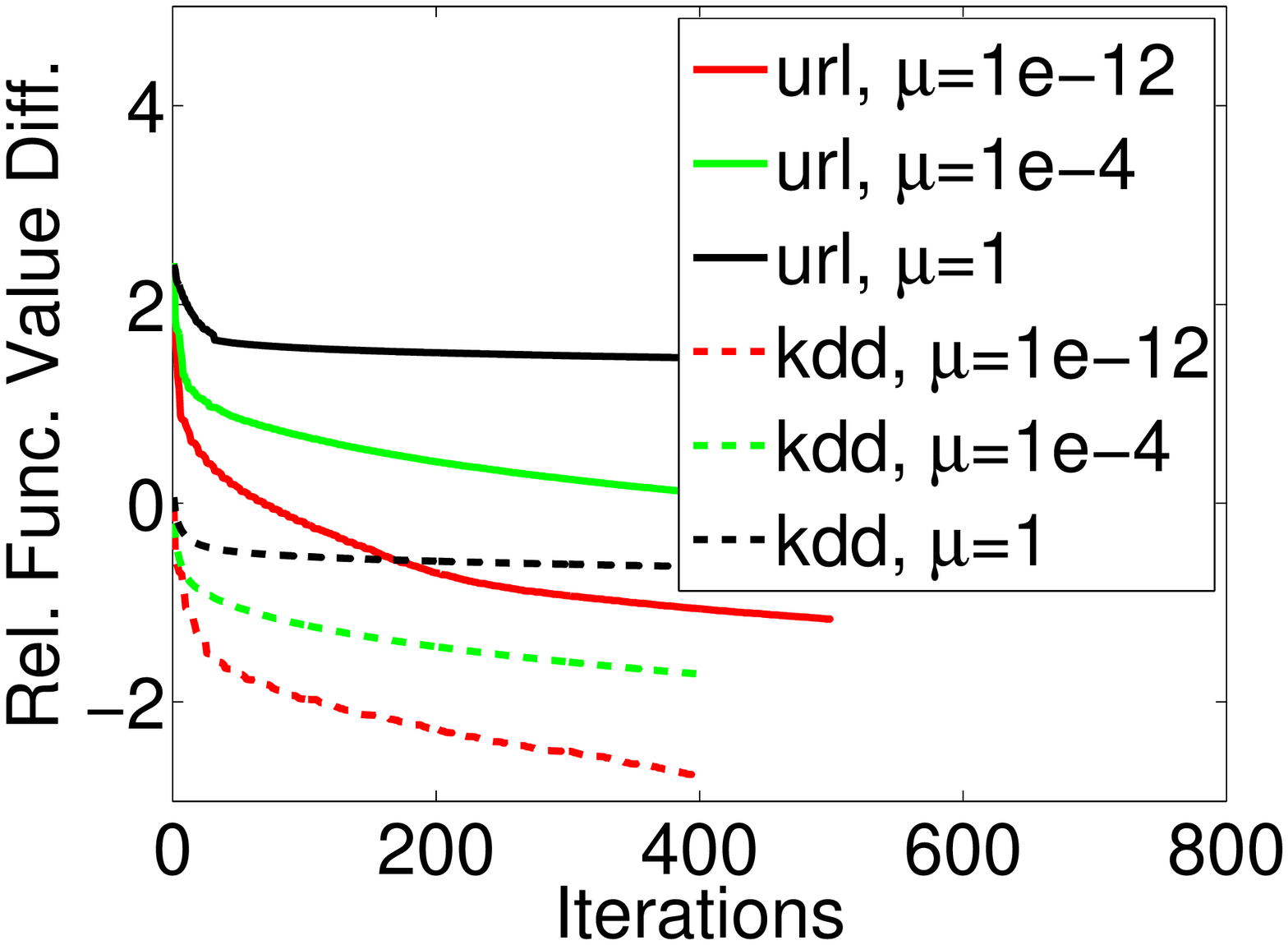}
\includegraphics[width=0.48\linewidth]{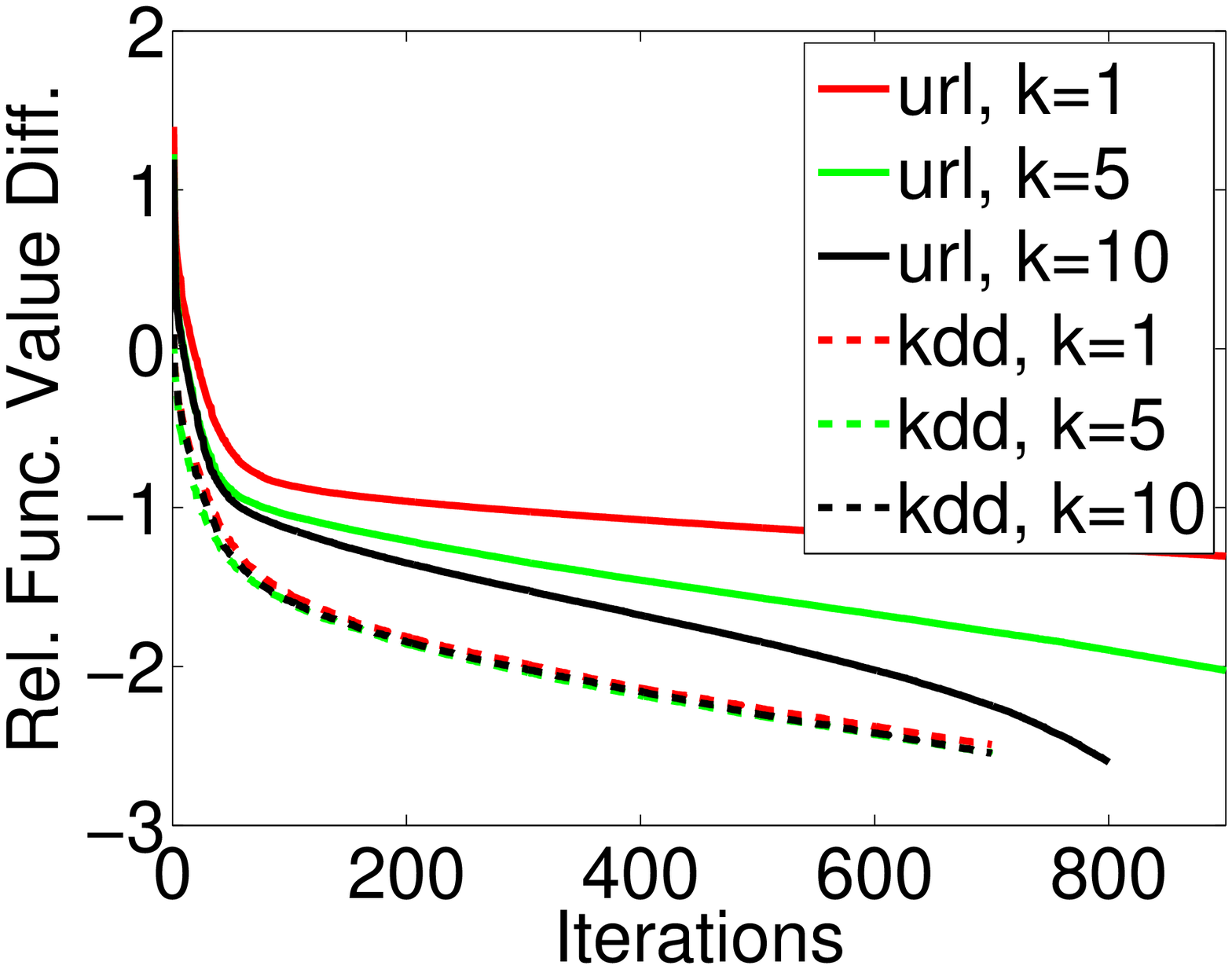}
\caption{\small{Left: the effect of $\mu$. Right: the effect of $k$, the number of cycles to minimize $f^t_p$. $\mu = 10^{-12}$ and $k = 10$ are good choices. $P=100$.}}
\label{fig:MuPlot}
\vspace{-0.15in}
\end{figure}

\begin{figure}[t]
\hspace{-0in}
\includegraphics[width=0.48\linewidth]{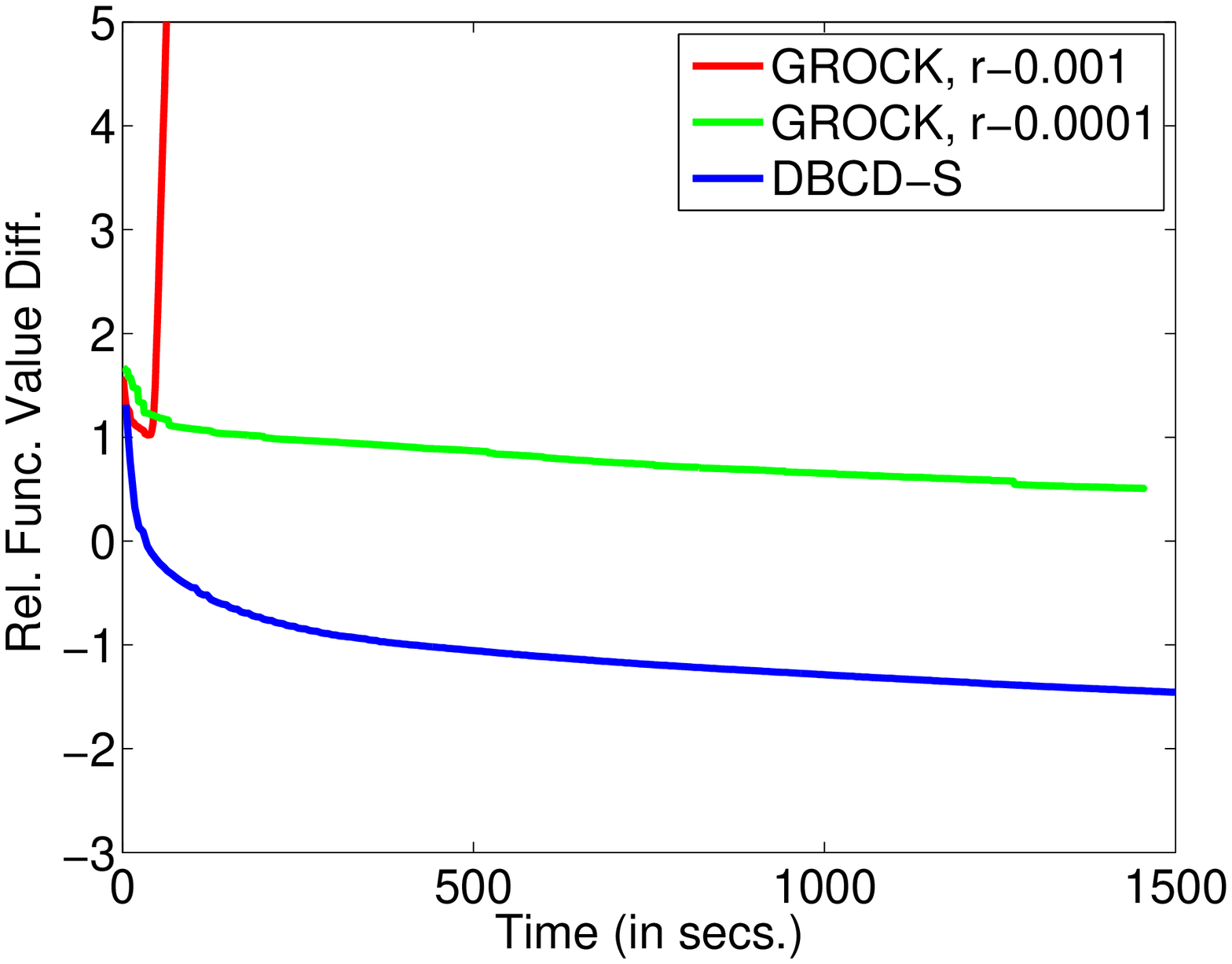}
\includegraphics[width=0.48\linewidth]{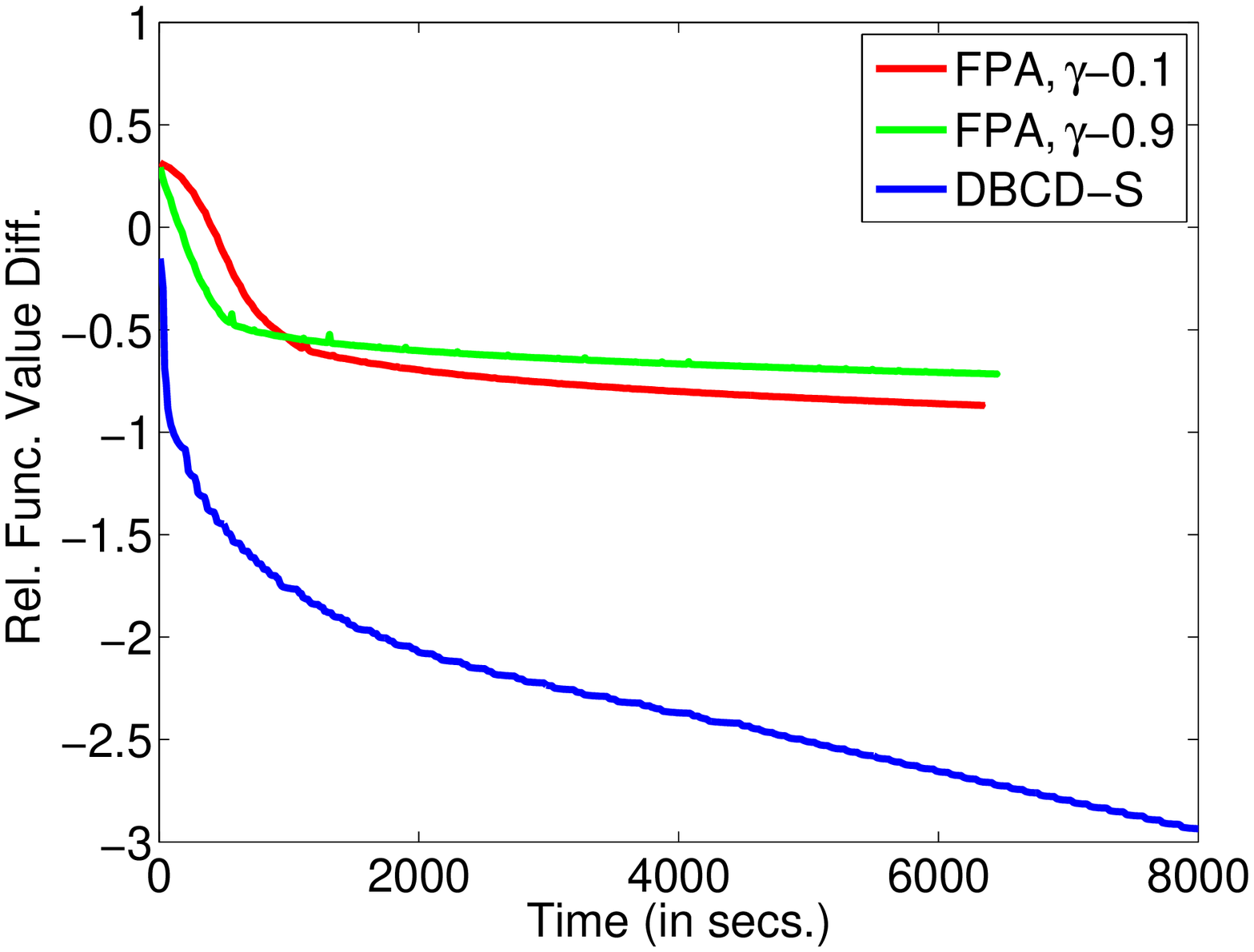}
\caption{\small{Left: Divergence and slow convergence of \grock~on the \textsc{URL} dataset ($\lambda = 2.4 \times 10^{-6}$ and $P = 25$). Right: Extremely slow convergence of \fpa~on the \textsc{KDD} dataset ($\lambda = 4.6 \times 10^{-7}$ and $P = 100$).}}
\label{fig:Issues}
\vspace{-0.15in}
\end{figure}

\subsection{Performance Evaluation}
\label{subsec:eval}

We begin by comparing the efficiency of various methods and demonstrating the superiority of the new methods that were motivated in Section~\ref{sec:motiv} and developed in Section~\ref{sec:dbcd}. After this we analyze and explain the reasons for the superiority. \vspace*{0.05in}

\noindent{\bf Study on AUPRC and RFVD:} 
We compare the performance of all methods by studying the variation of AUPRC and RFVD as a function of time, for various choices of $\lambda$, working set size (WSS) and the number of nodes ($P$) on \textsc{KDD} and \textsc{URL} datasets. To avoid cluttering with too many plots, we provide only representative ones; but, the observations that we make below hold for others too.

Figure~\ref{fig:kddobj} shows the objective function plots for (\textit{KDD}) with $\lambda$ set to $4.6 \times 10^{-7}$. We see that \dbcds~clearly outperforms all other methods; for example, if we set the RFVD value to $-2$ as the stopping criterion, \dbcds~is faster than existing methods by an order of magnitude. \pcdns~comes as the second best. The S-scheme gives significant speed improvement over the R-scheme.
As we compare the performance for two different WSS (see Figure~\ref{fig:kddobj}(a)(b)), larger WSS gives some improvement and this speed-up is significant for \richtarik, \pcdnr~and \dbcdr. Note that \admm~is WSS independent since all the variables are updated. Because all variables are updated, ADMM performs slightly better than \richtarik, \pcdnr~and \dbcdr~when WSS is small (see Figure~\ref{fig:kddobj}(a)(c)). In this case, other methods take some time to reach optimal values, when the working set is selected randomly using the R-scheme. If we compare \admm~and \dbcds, \admm~is inferior; this holds even if we leave out the initial time needed for setting $\rho$.

\begin{figure*}[t]
\centering
\subfigure[$P = 25$, $WSS = 8086$]{
\includegraphics[width=0.45\linewidth]{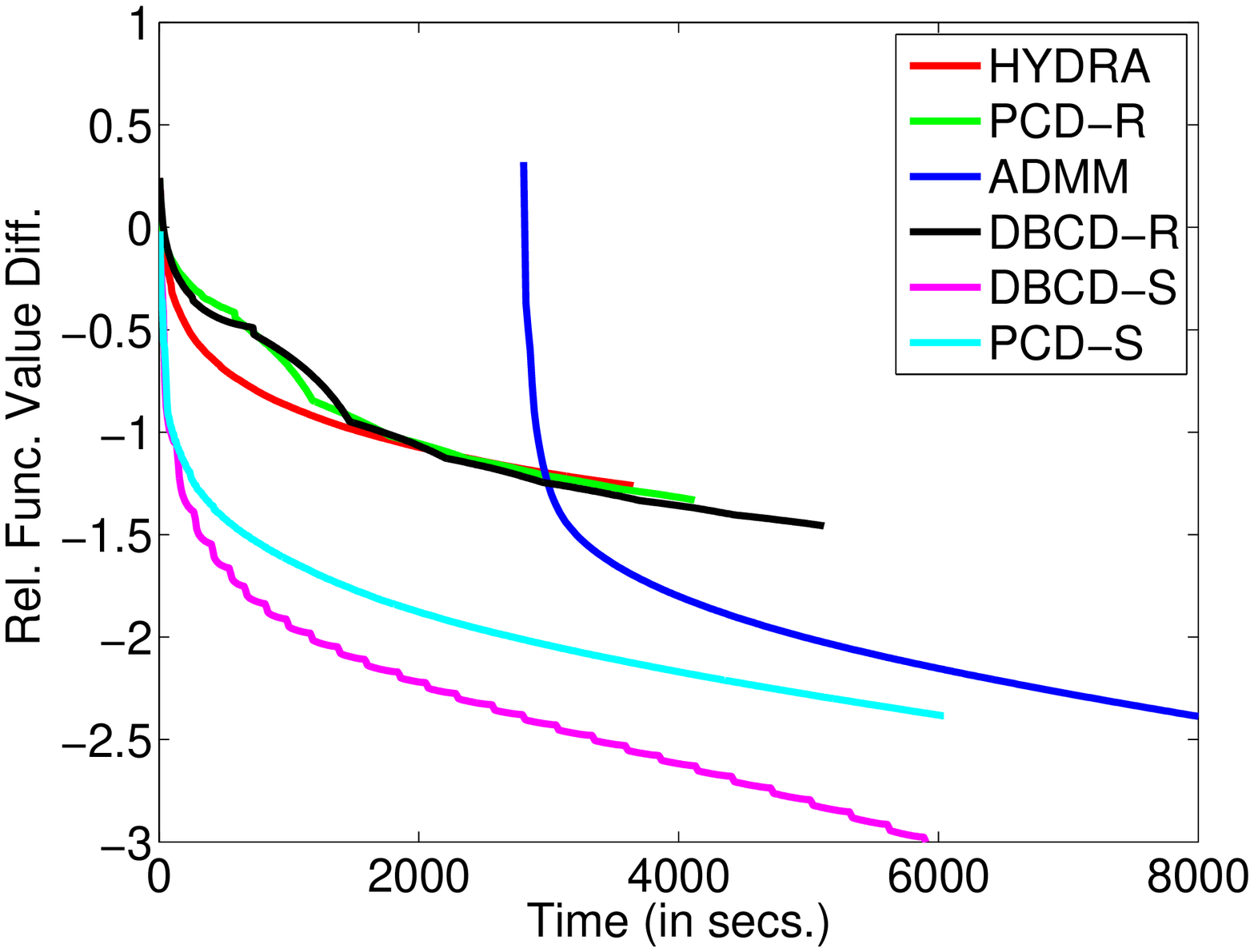}
\vspace*{-2in}
}
\subfigure[$P = 25$, $WSS = 80867$]{
\includegraphics[width=0.45\linewidth]{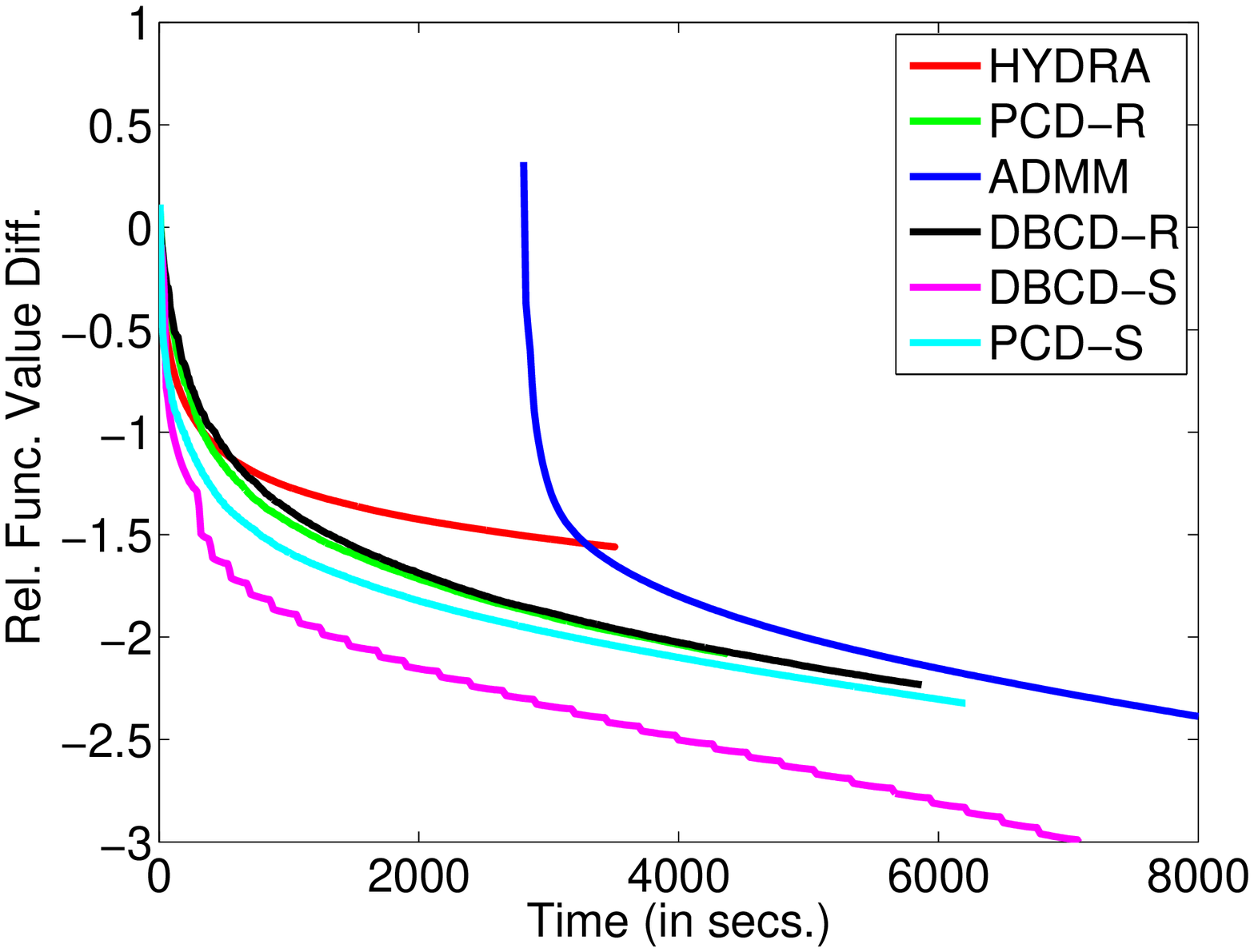}
\vspace*{-2in}
}
\subfigure[$P = 100$, $WSS = 2021$]{
\includegraphics[width=0.45\linewidth]{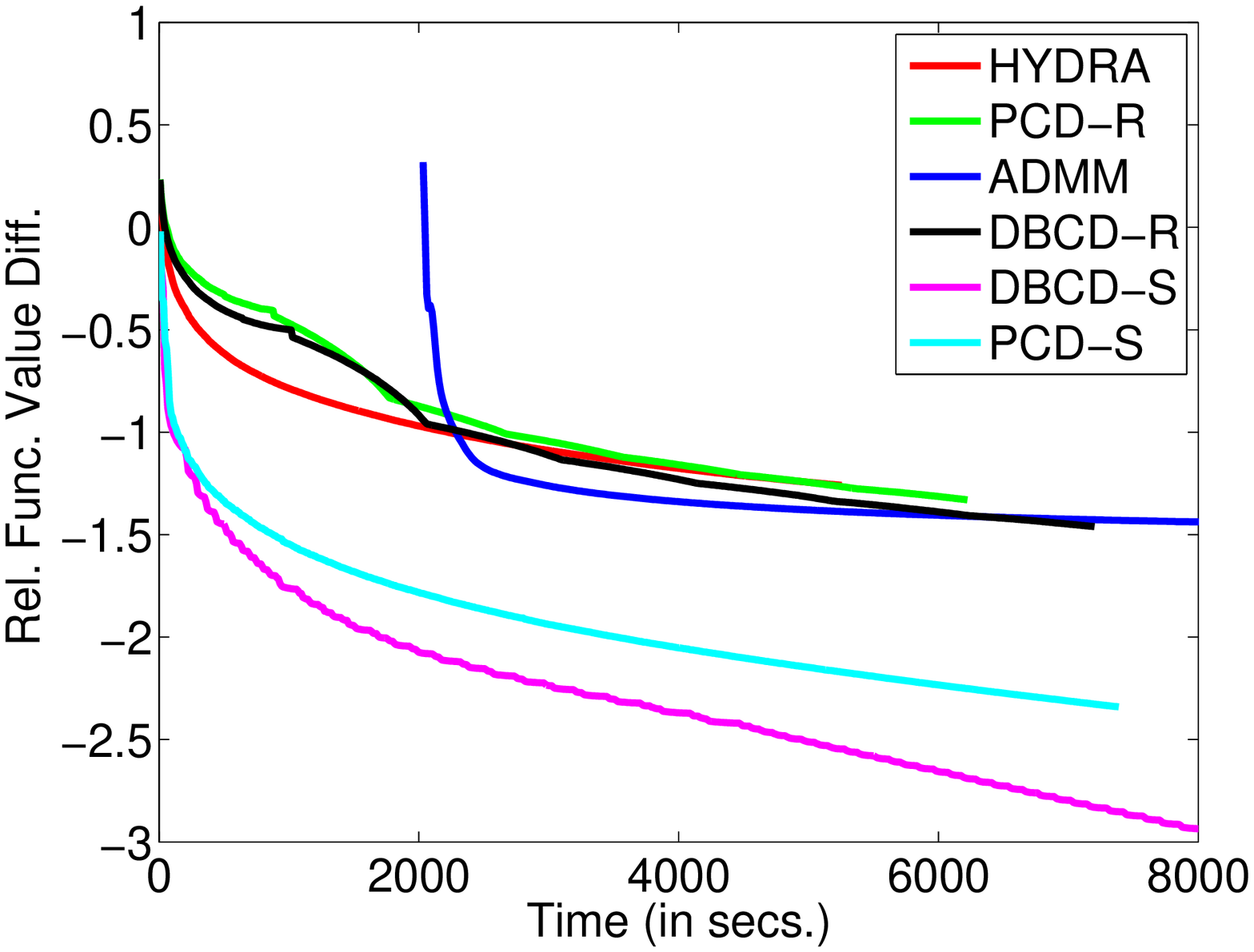}
\vspace*{-2in}
}
\subfigure[$P = 100$, $WSS = 20216$]{
\includegraphics[width=0.45\linewidth]{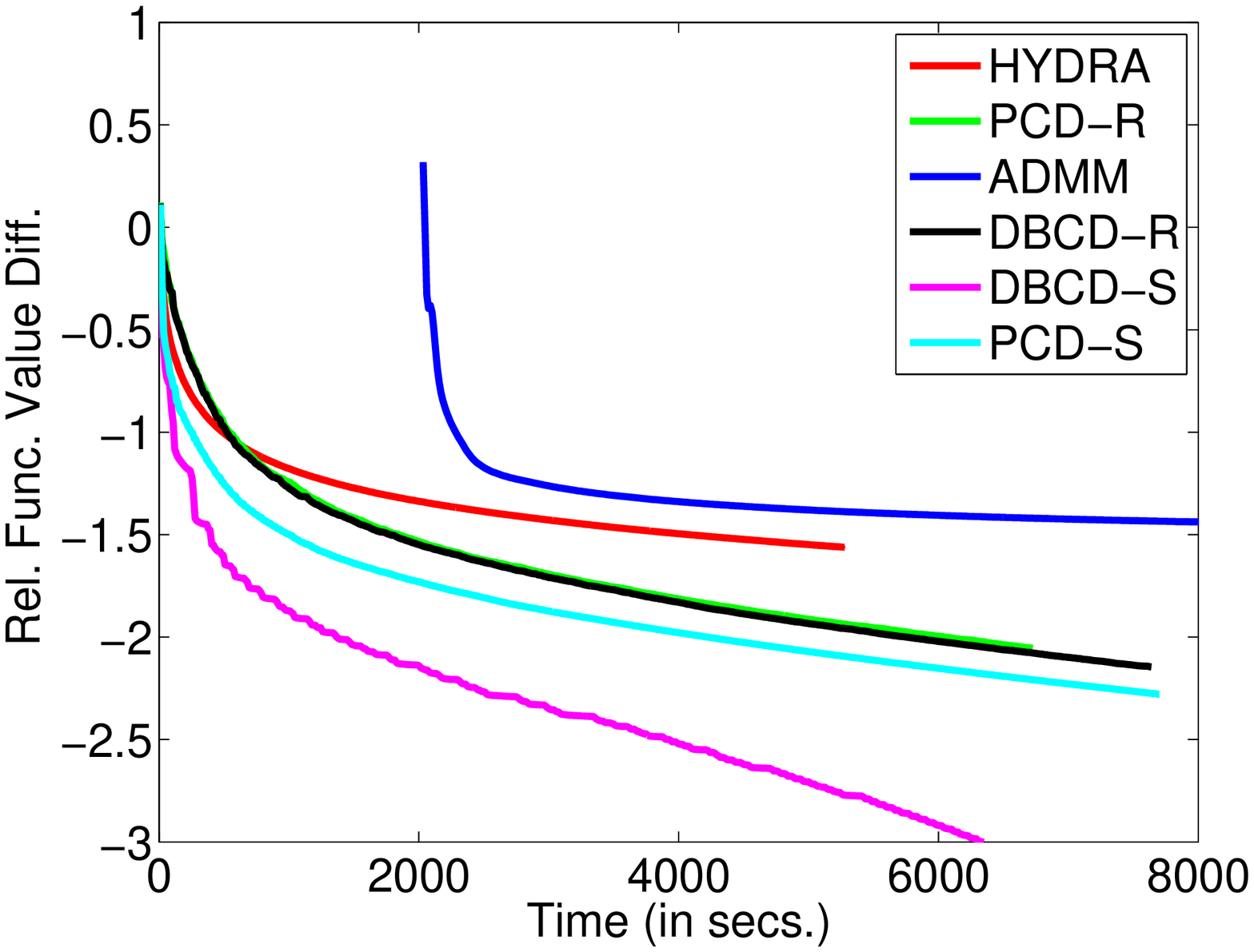}
\vspace*{-2in}
}
\caption{\small{\textsc{KDD} dataset. Relative function value difference in log scale. $\lambda = 4.6 \times 10^{-7}$}}
\label{fig:kddobj}
\vspace{-0.15in}
\end{figure*}

\begin{figure*}[t]
\centering
\subfigure[$P = 25$, $WSS = 1292$]{
\includegraphics[width=0.45\linewidth]{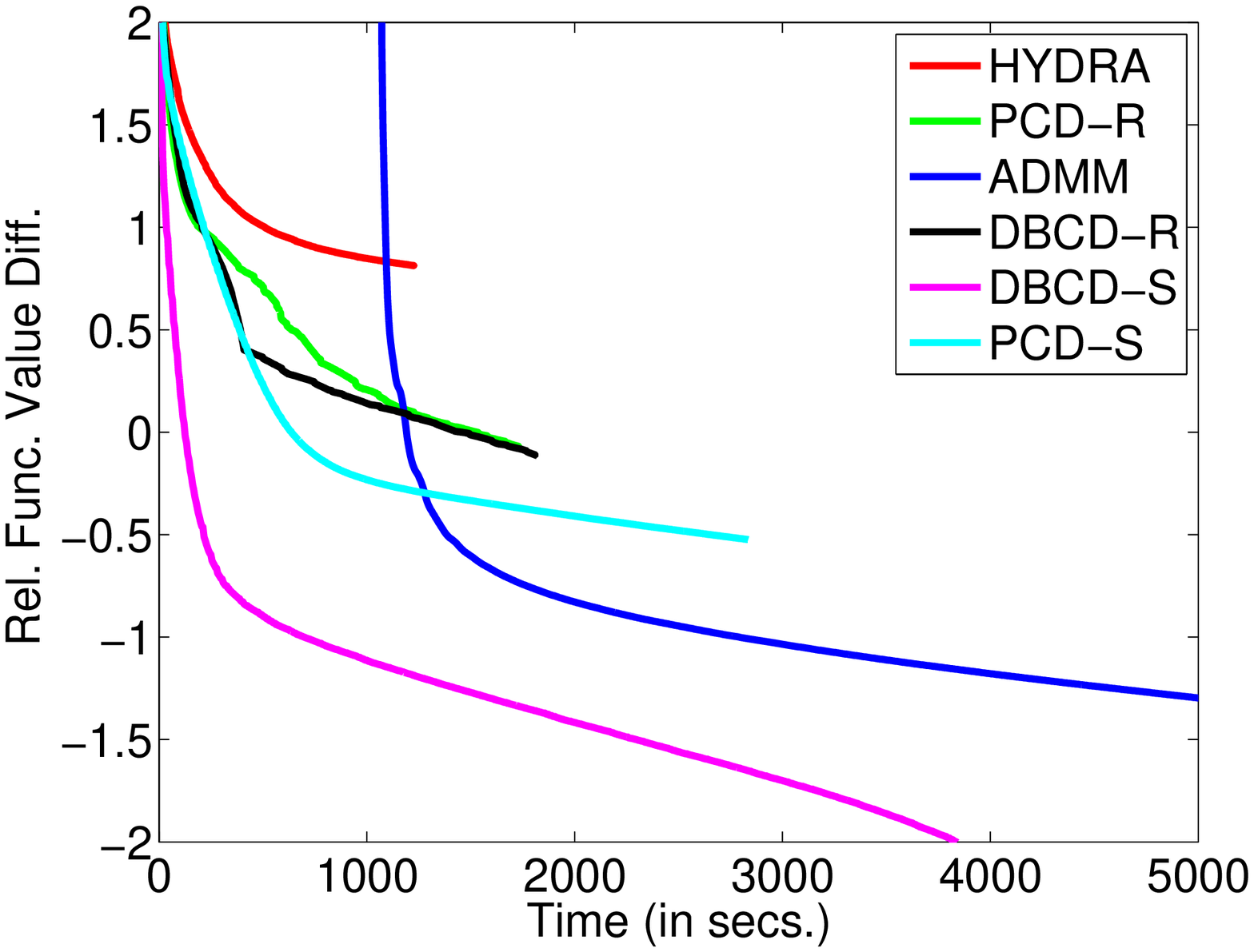}
\vspace*{-2in}
}
\subfigure[$P = 25$, $WSS = 12927$]{
\includegraphics[width=0.45\linewidth]{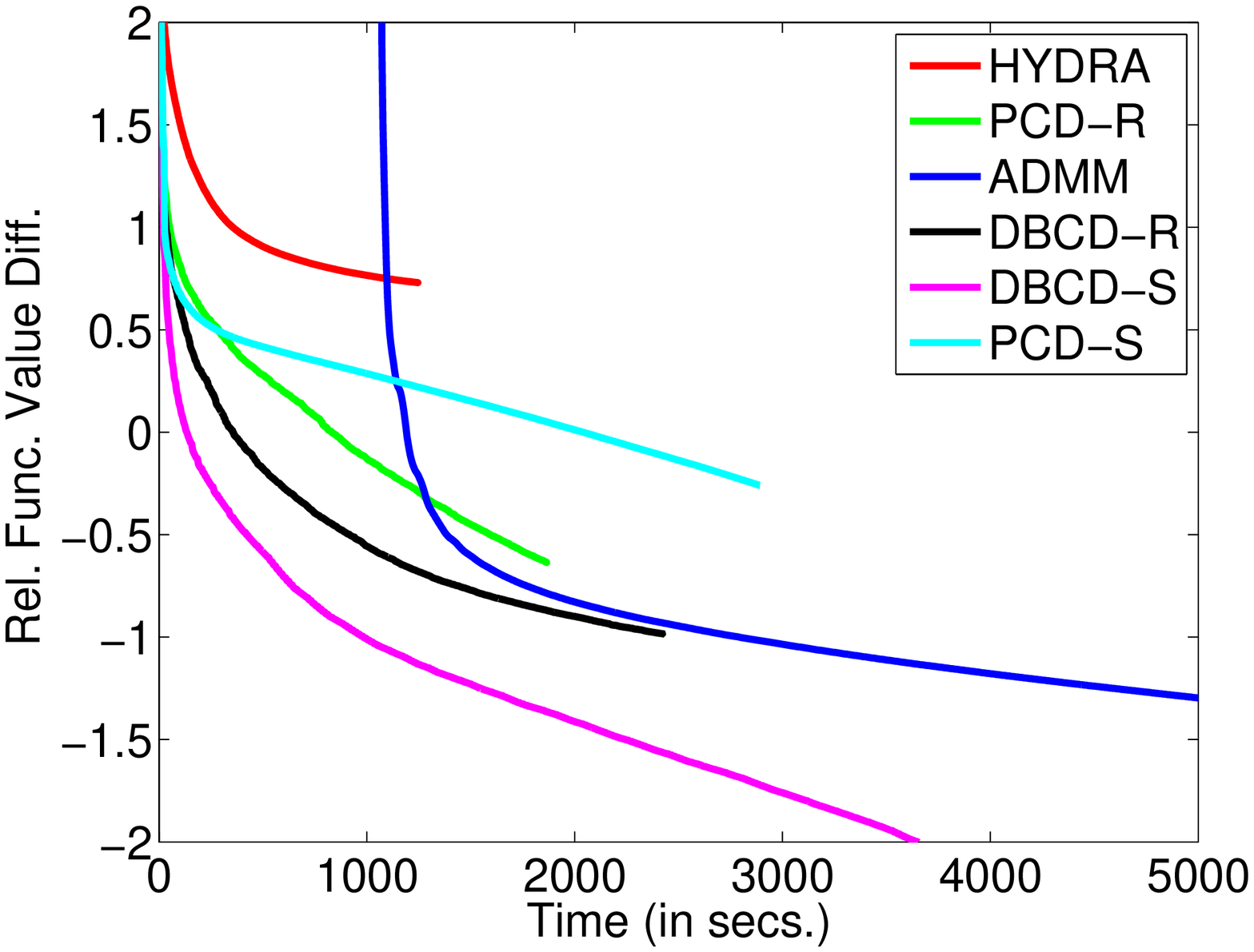}
\vspace*{-2in}
}
\subfigure[$P = 100$, $WSS = 323$]{
\includegraphics[width=0.45\linewidth]{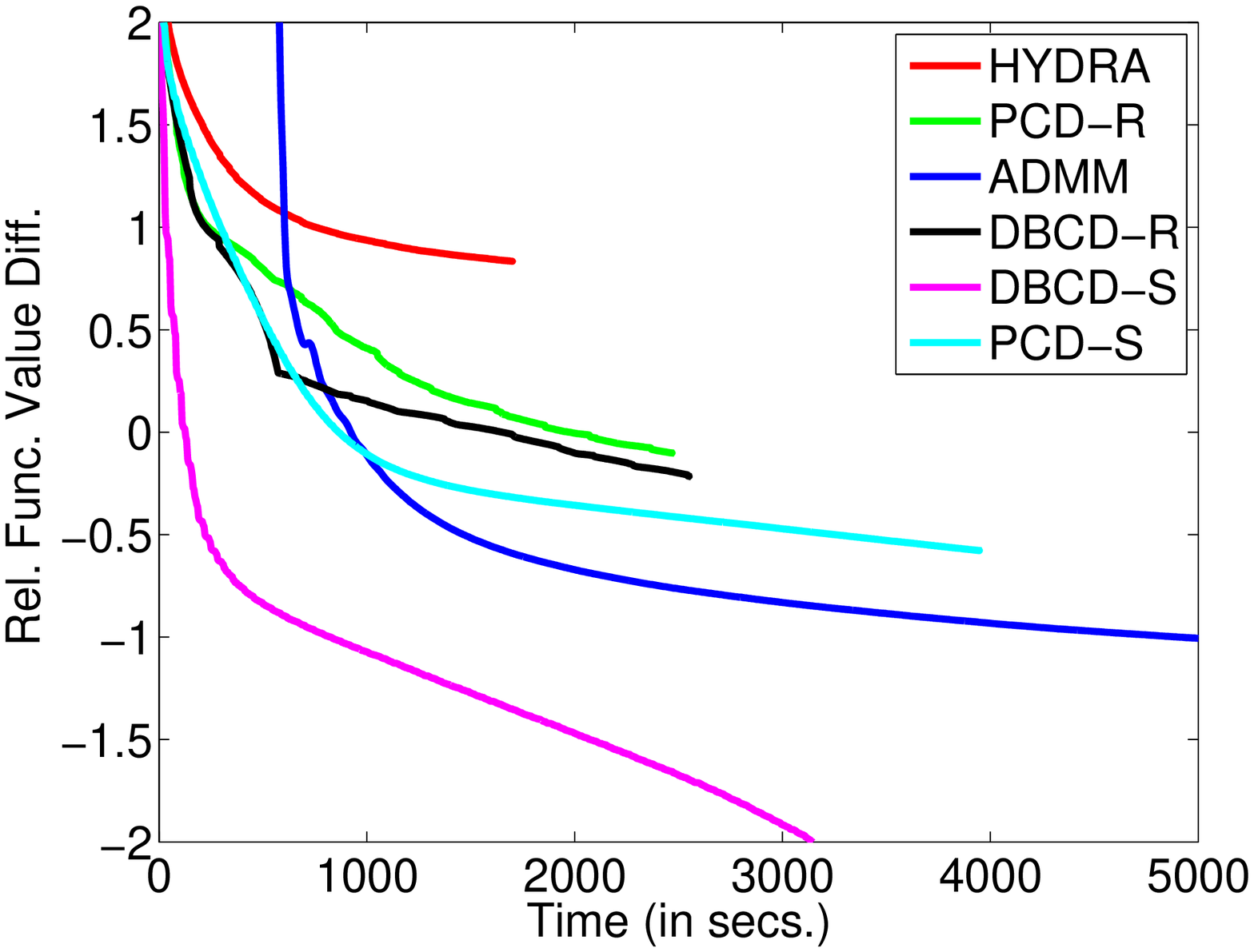}
\vspace*{-2in}
}
\subfigure[$P = 100$, $WSS = 3231$]{
\includegraphics[width=0.45\linewidth]{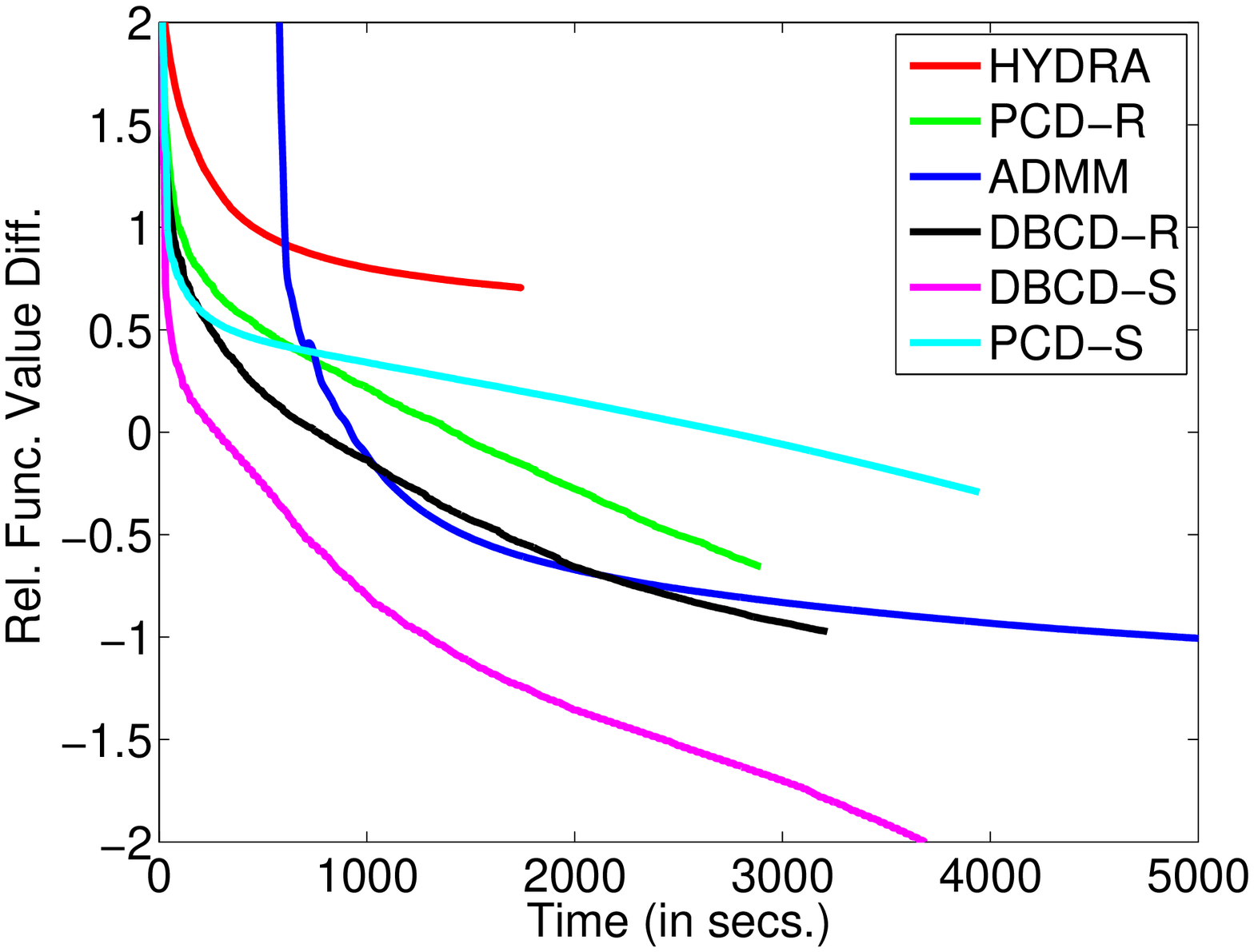}
\vspace*{-2in}
}
\caption{\small{\textsc{URL} dataset. Relative function value difference in log scale. $\lambda = 9.0 \times 10^{-8}$}}
\label{fig:urlobj}
\vspace{-0.15in}
\end{figure*}

\begin{figure*}[t]
\centering
\subfigure[$P = 25$, $WSS = 8086$]{
\includegraphics[width=0.45\linewidth]{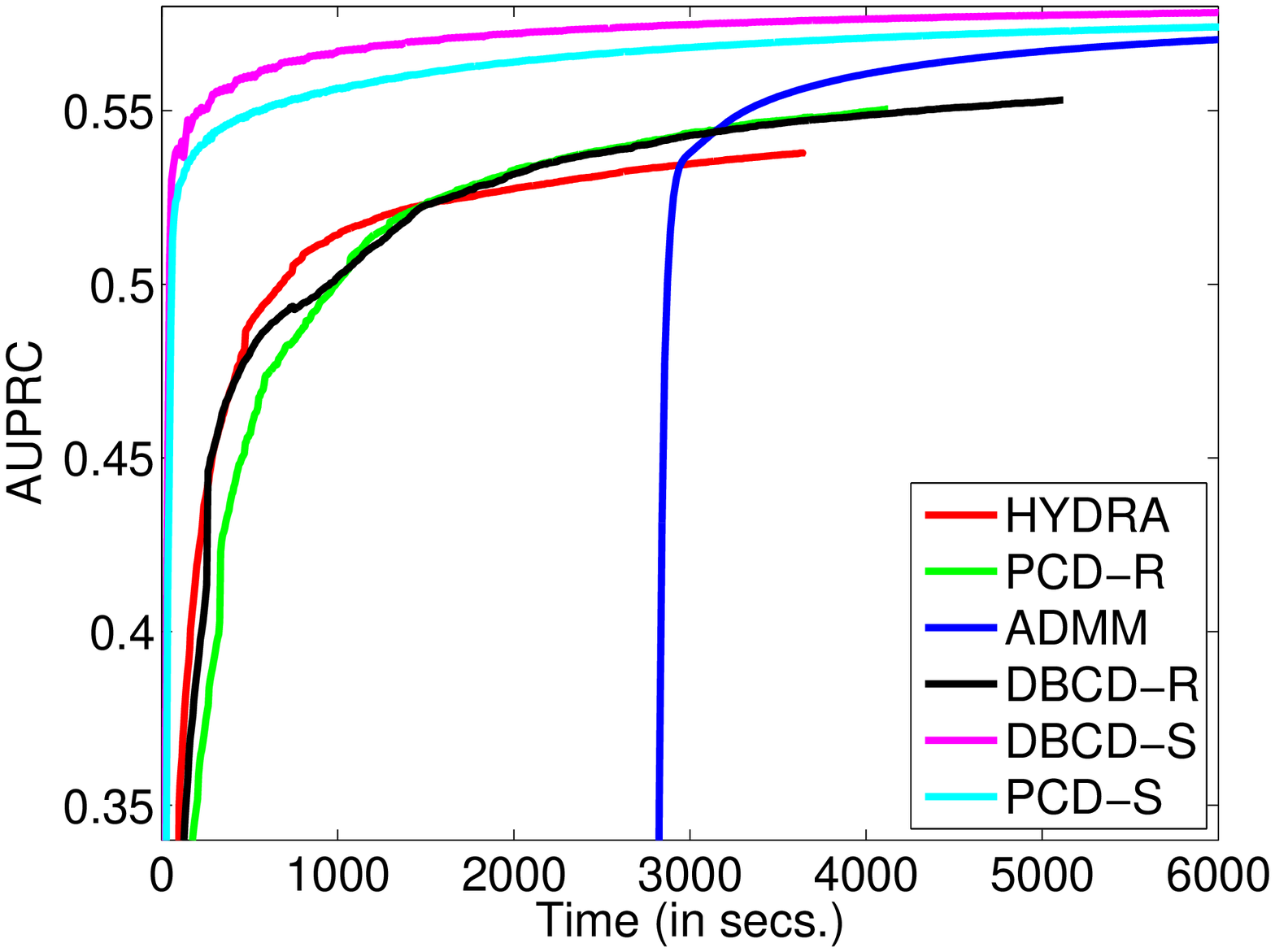}
}
\subfigure[$P = 25$, $WSS = 80867$]{
\includegraphics[width=0.45\linewidth]{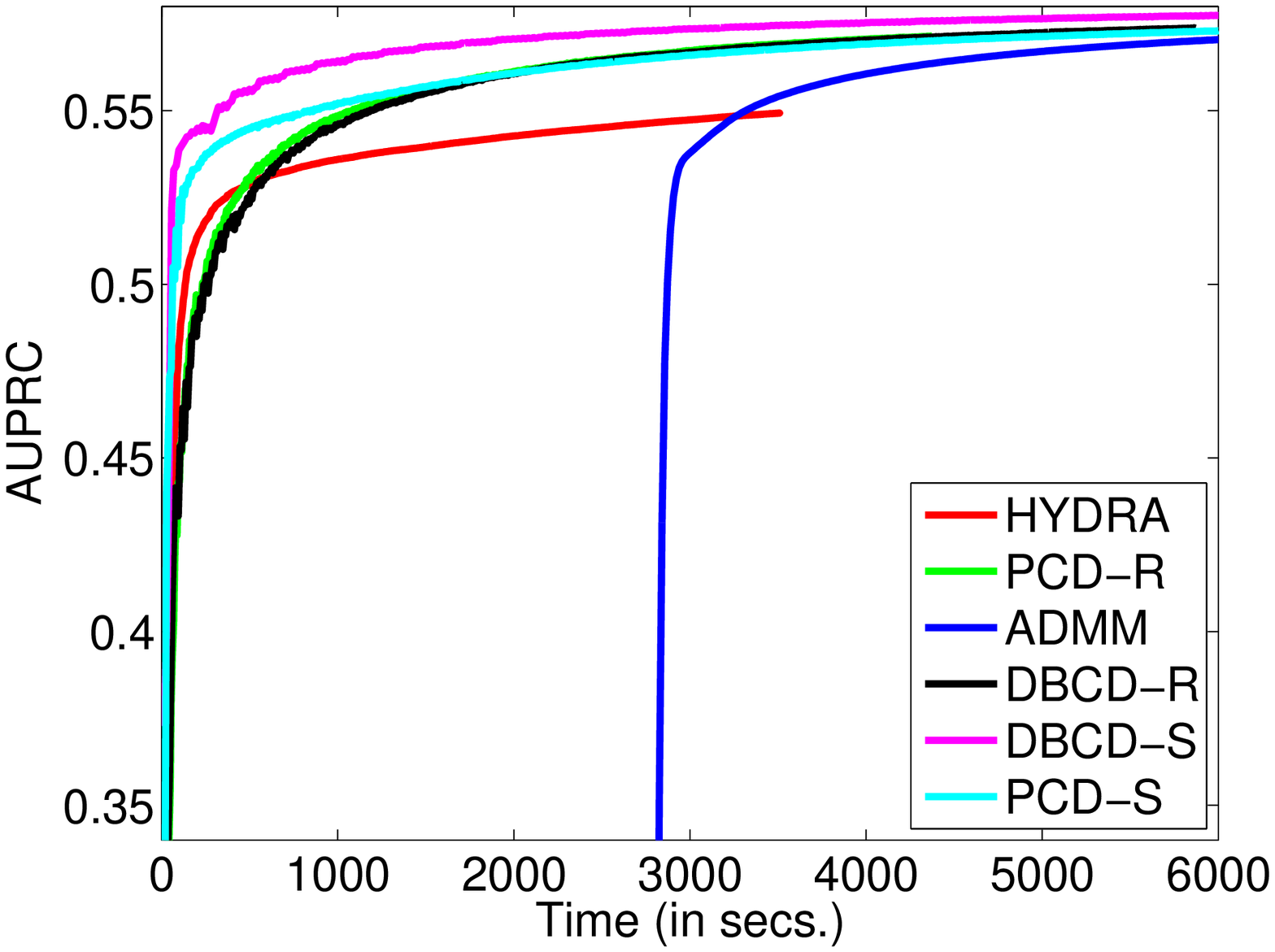}
}
\subfigure[$P = 100$, $WSS = 2021$]{
\includegraphics[width=0.45\linewidth]{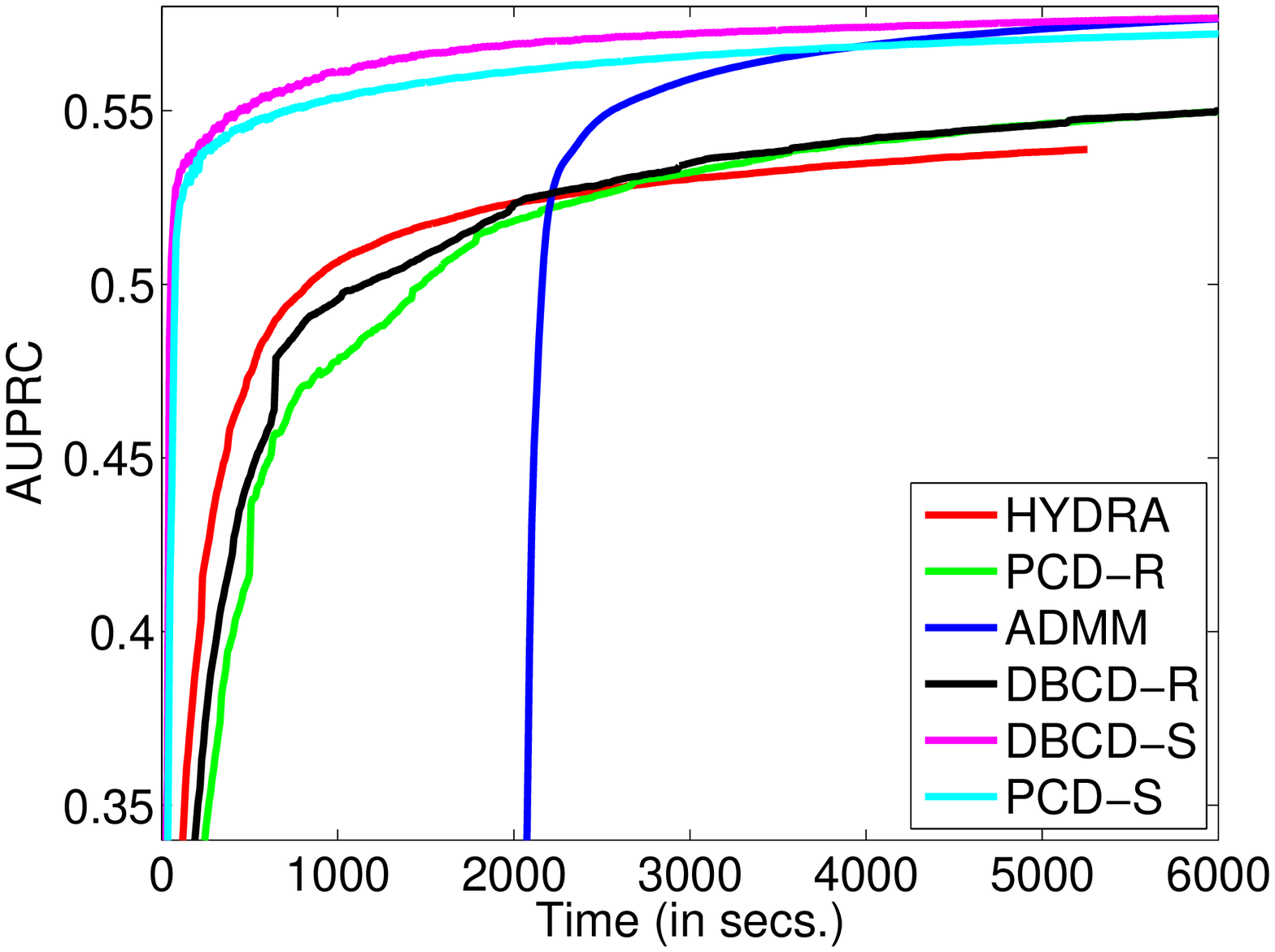}
}
\subfigure[$P = 100$, $WSS = 20216$]{
\includegraphics[width=0.45\linewidth]{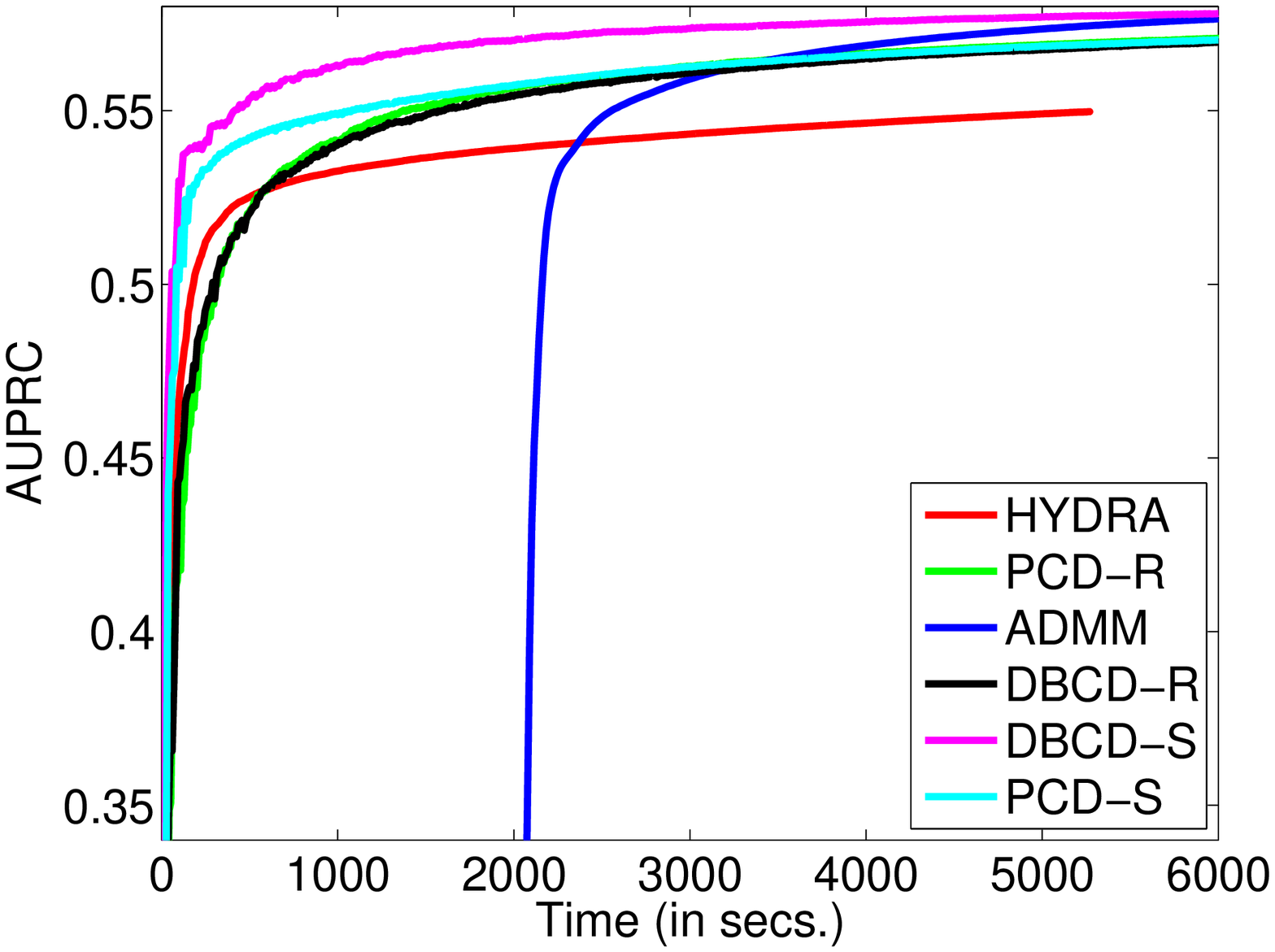}
}
\caption{\small{\textsc{KDD} dataset. AUPRC Plots. $\lambda = 4.6 \times 10^{-7}$}}
\label{fig:kddauprc}
\end{figure*}

\begin{figure*}[t]
\centering
\subfigure[$P = 25$, $WSS = 1292$]{
\includegraphics[width=0.45\linewidth]{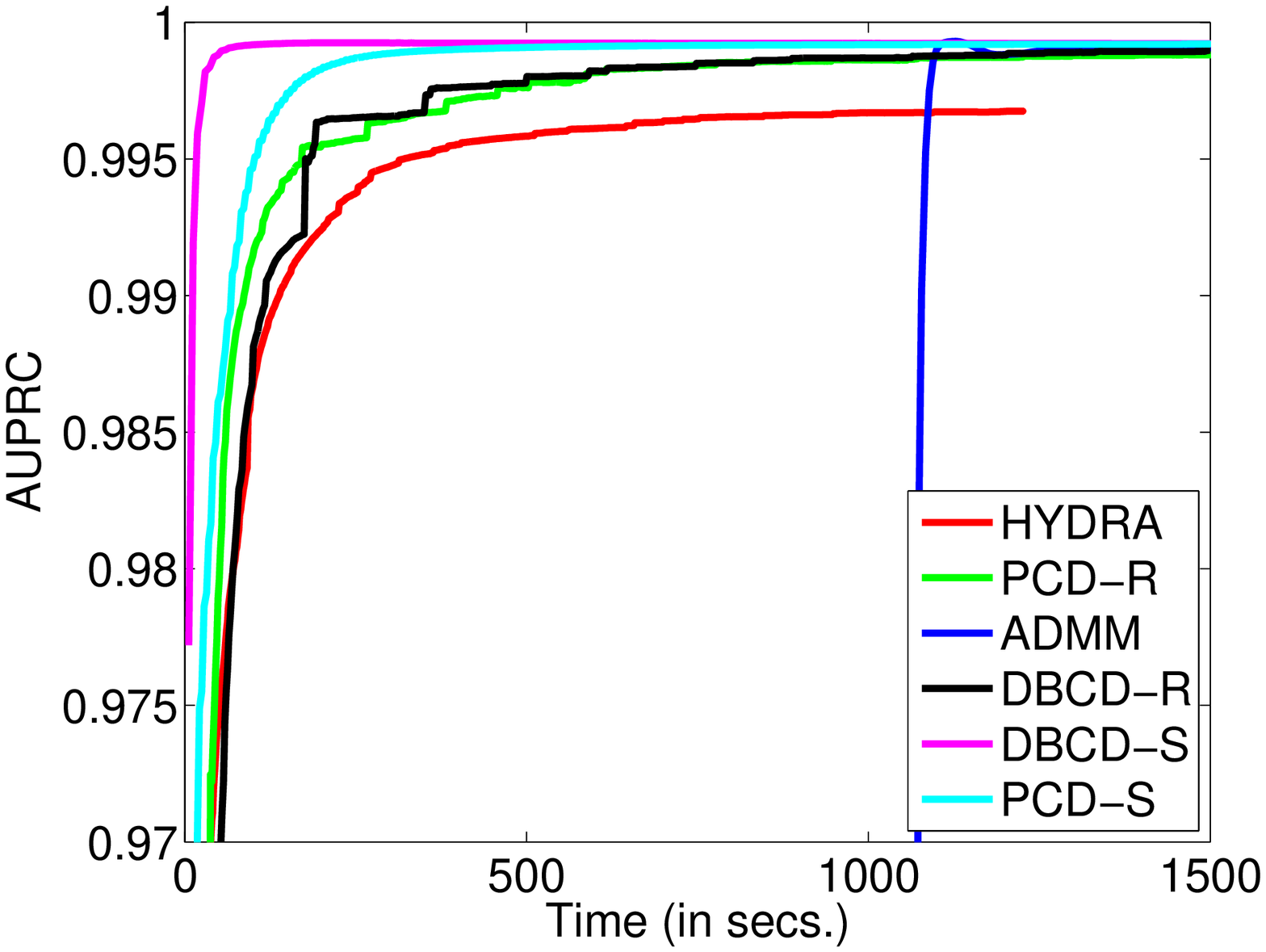}
}
\subfigure[$P = 25$, $WSS = 12927$]{
\includegraphics[width=0.45\linewidth]{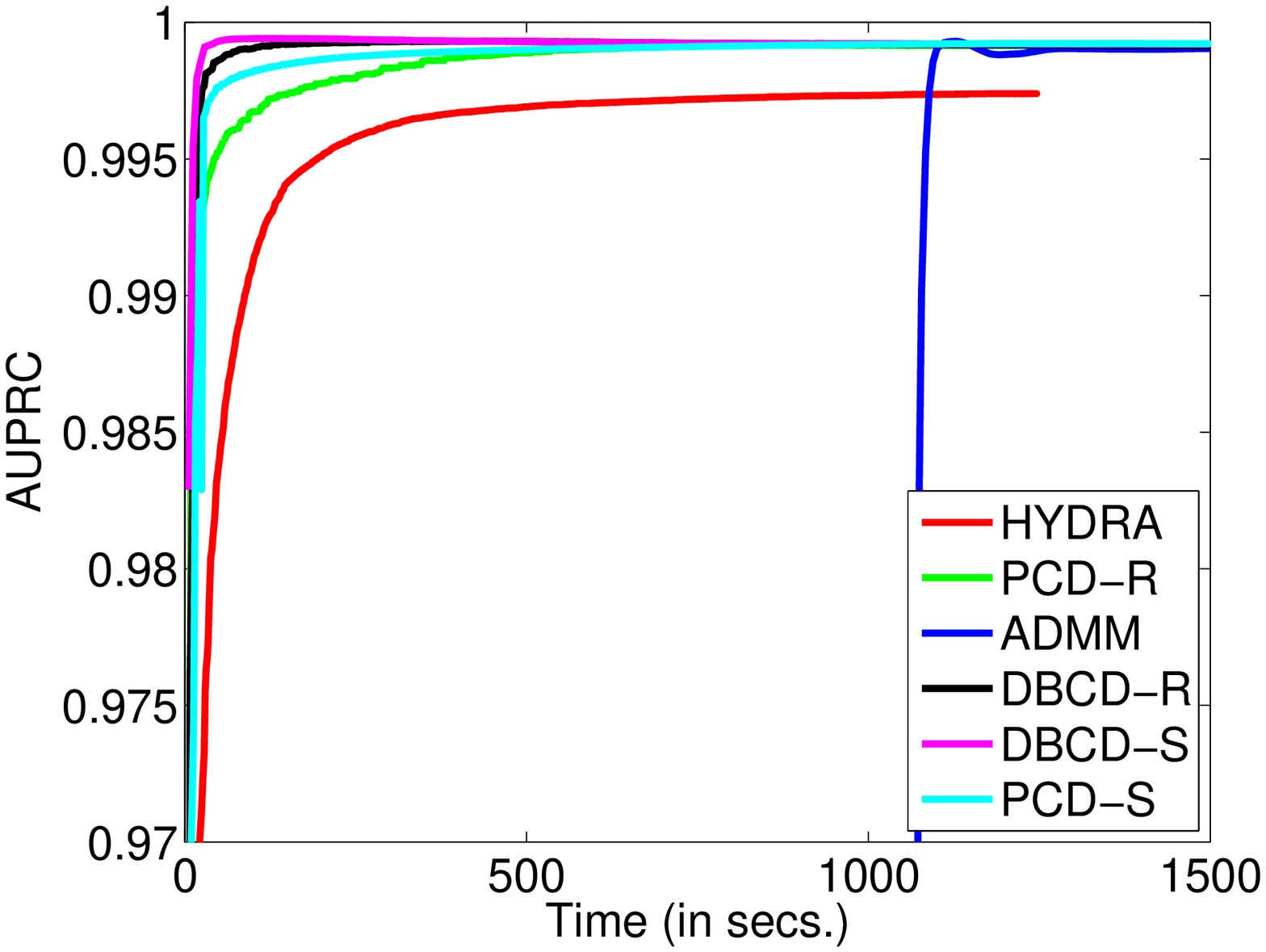}
}
\subfigure[$P = 100$, $WSS = 323$]{
\includegraphics[width=0.45\linewidth]{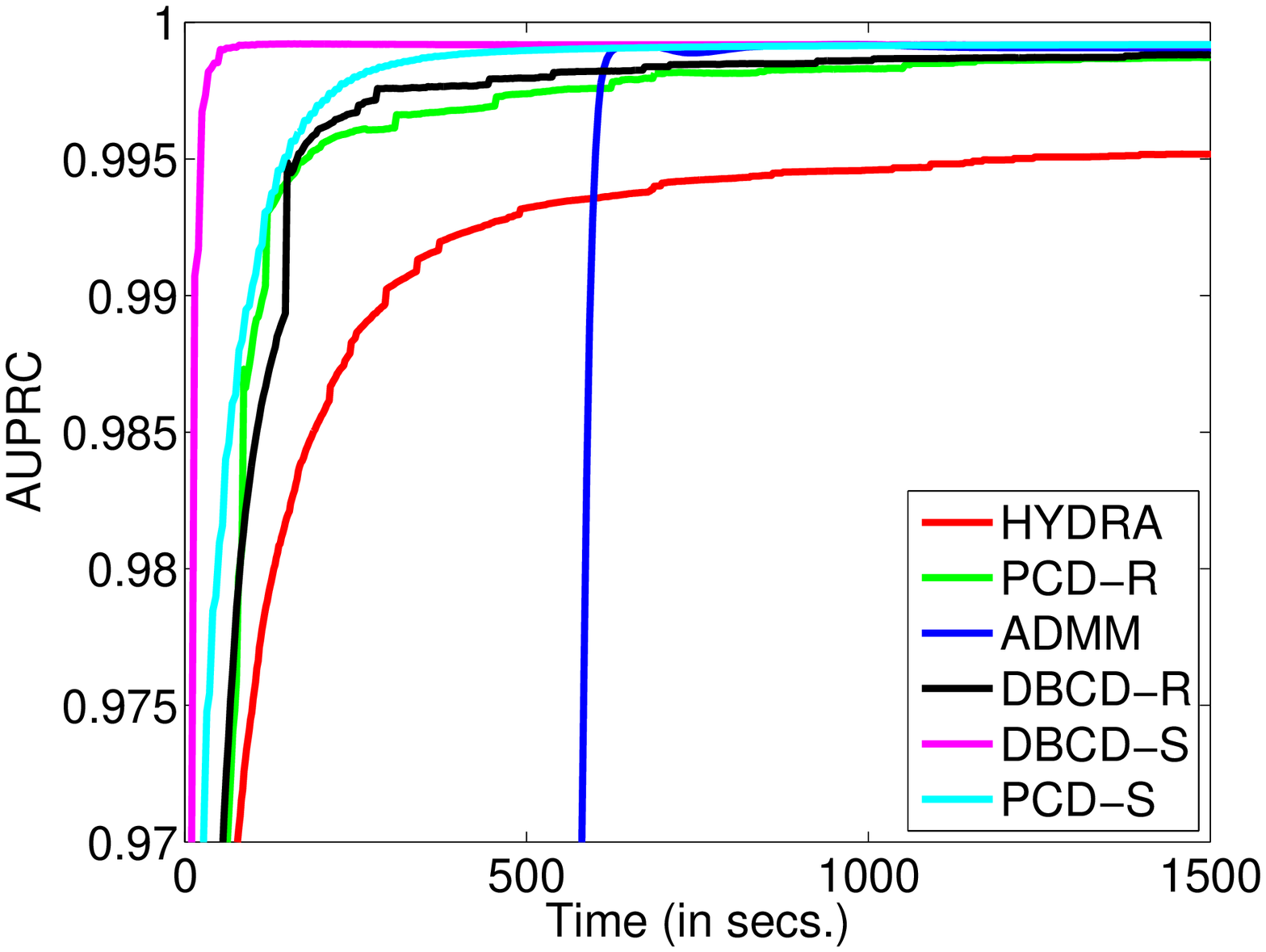}
}
\subfigure[$P = 100$, $WSS = 3231$]{
\includegraphics[width=0.45\linewidth]{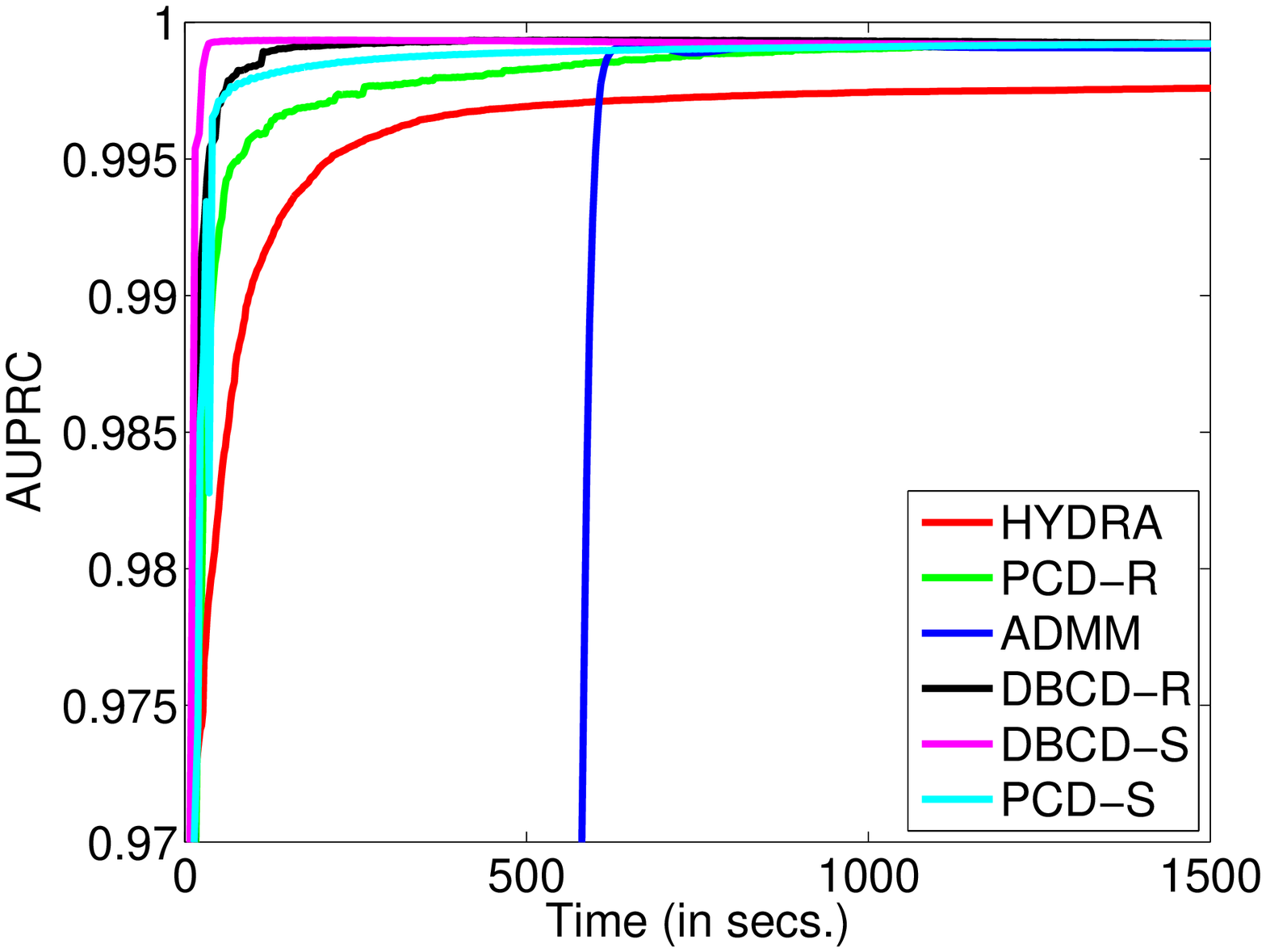}
}
\caption{\small{\textsc{URL} dataset. AUPRC plots. $\lambda = 9.0 \times 10^{-8}$}}
\label{fig:urlauprc}
\end{figure*}

\begin{figure*}[t]
\centering
\subfigure[\textsc{KDD}, $WSS = 2021$]{
\includegraphics[width=0.45\linewidth]{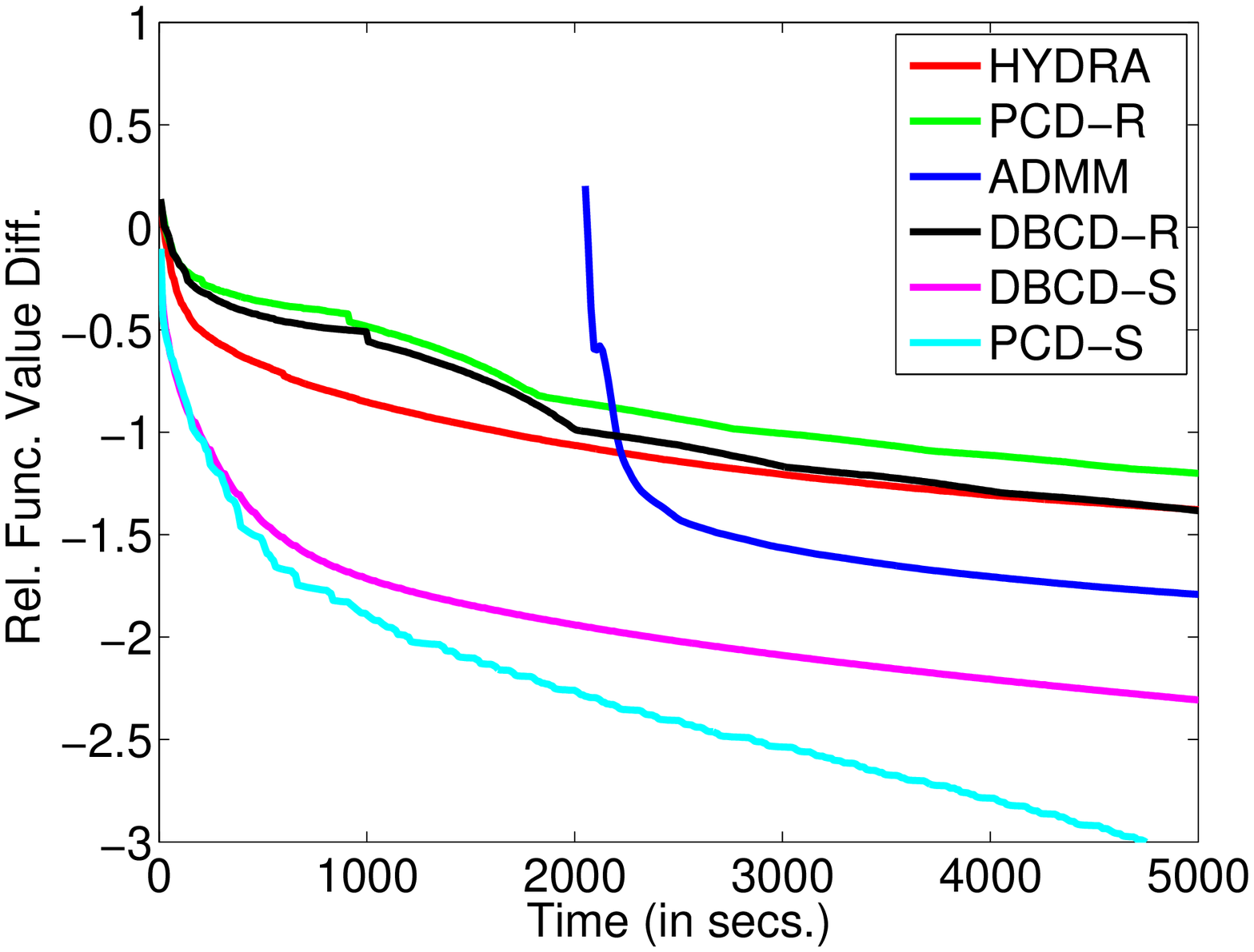}
}
\subfigure[\textsc{KDD}, $WSS = 20216$]{
\includegraphics[width=0.45\linewidth]{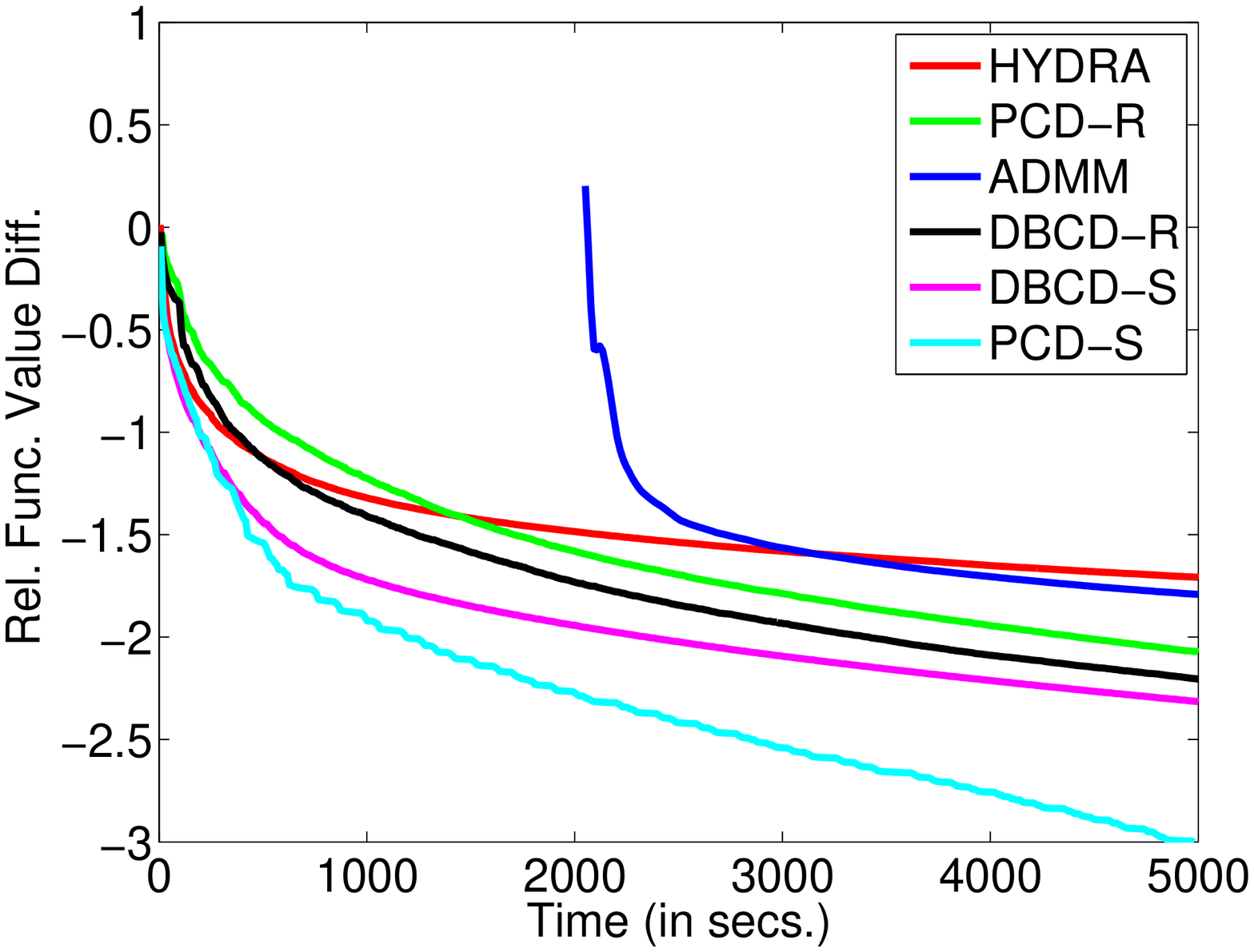}
}
\subfigure[\textsc{URL}, $WSS = 323$]{
\includegraphics[width=0.45\linewidth]{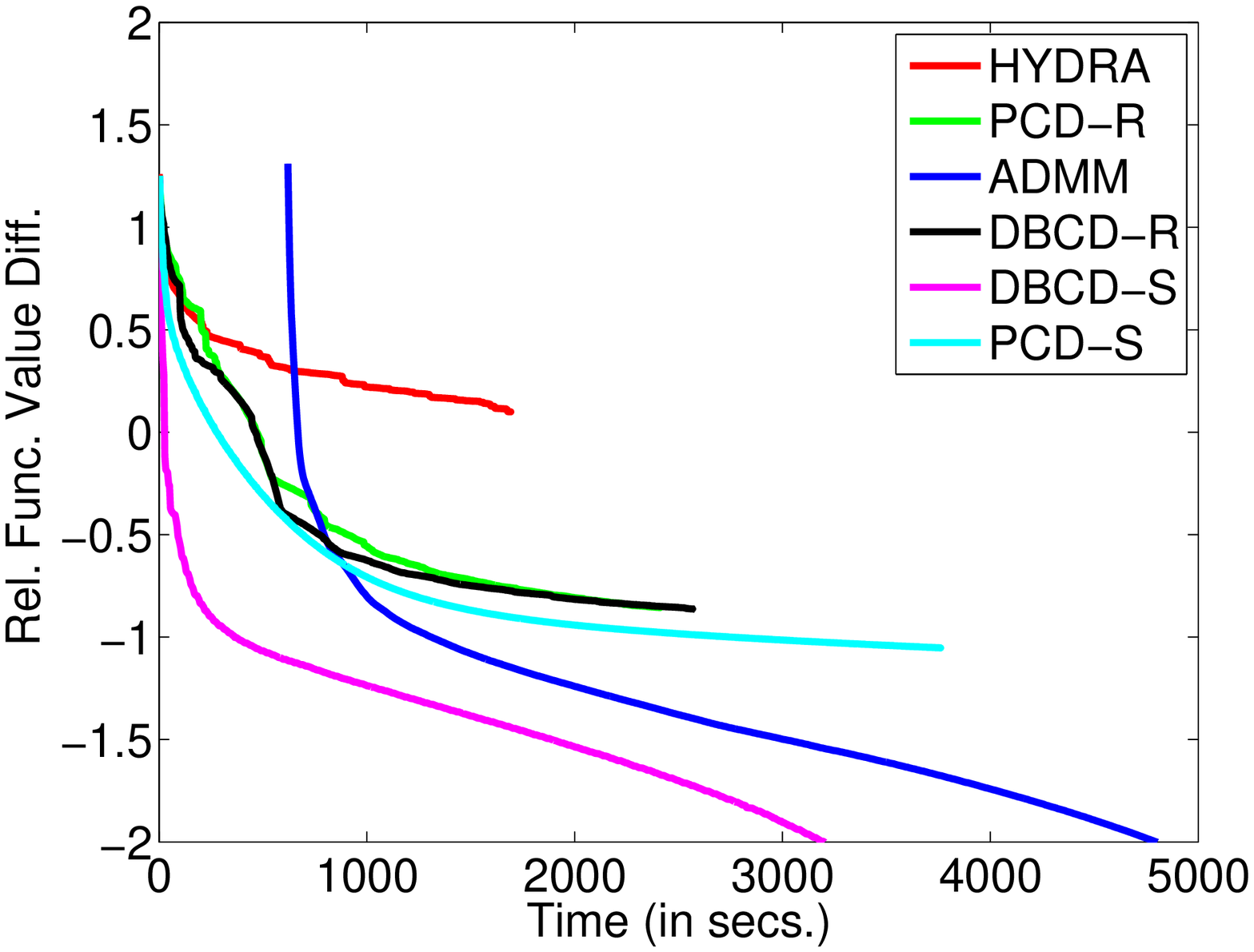}
}
\subfigure[\textsc{URL}, $WSS = 3231$]{
\includegraphics[width=0.45\linewidth]{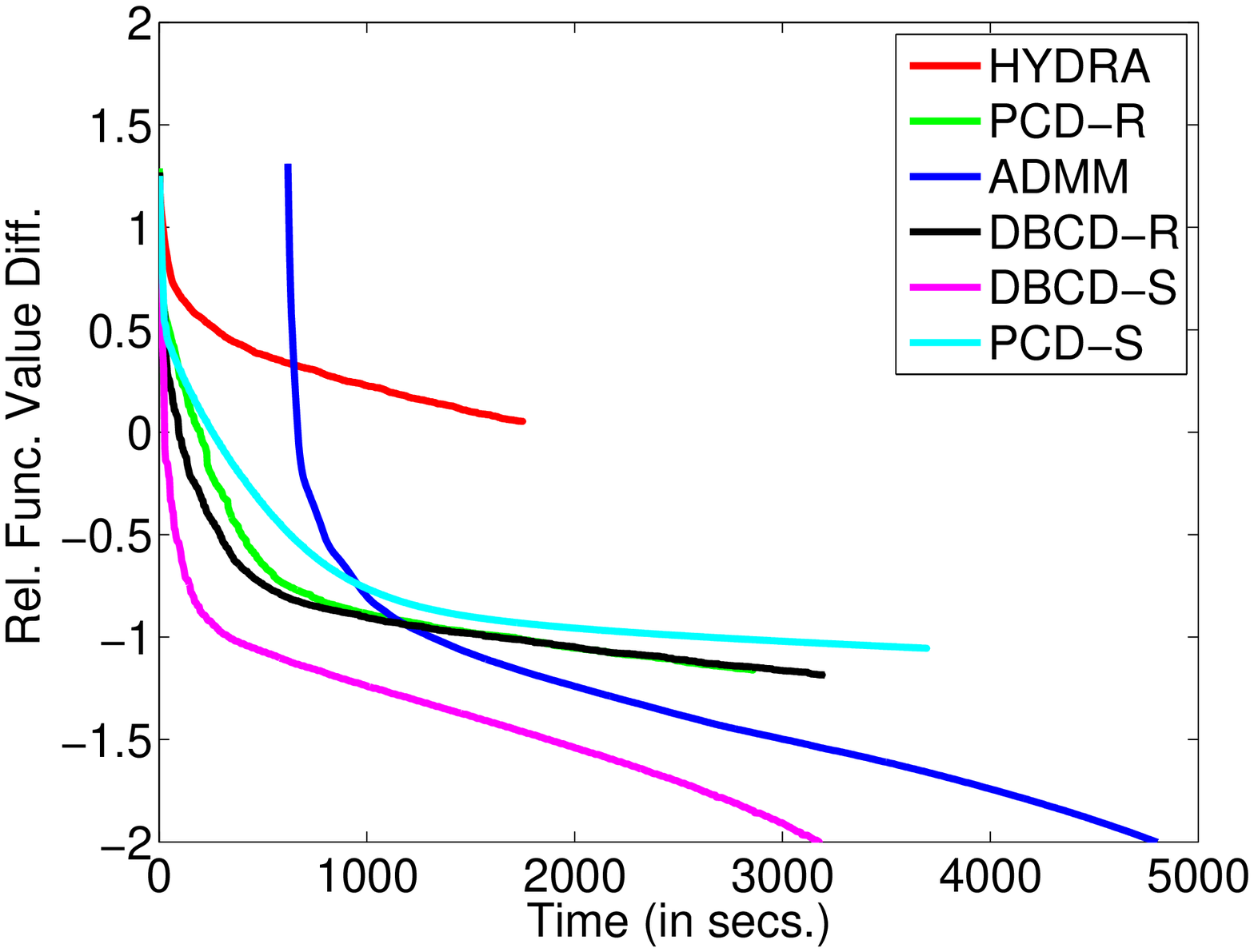}
}
\caption{\small{Relative function value difference in log scale. \textsc{KDD} dataset: $\lambda = 1.2 \times 10^{-5}$. \textsc{URL} dataset: $\lambda = 7.3 \times 10^{-7}$}}
\label{fig:kddurlobj}
\end{figure*}

\begin{figure*}[t]
\centering
\subfigure[\textsc{KDD}, $WSS = 2021$]{
\includegraphics[width=0.45\linewidth]{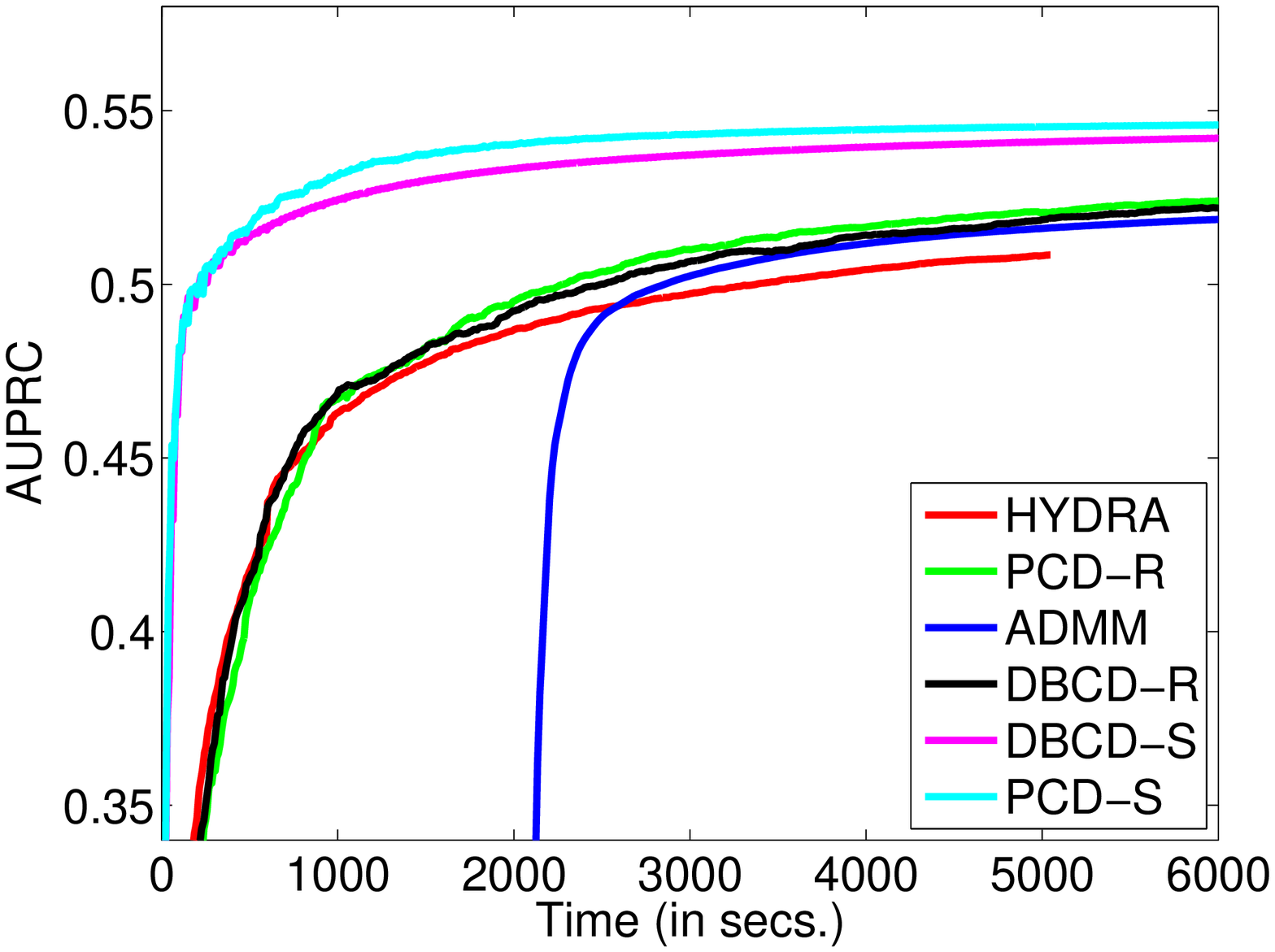}
}
\subfigure[\textsc{KDD}, $WSS = 20216$]{
\includegraphics[width=0.45\linewidth]{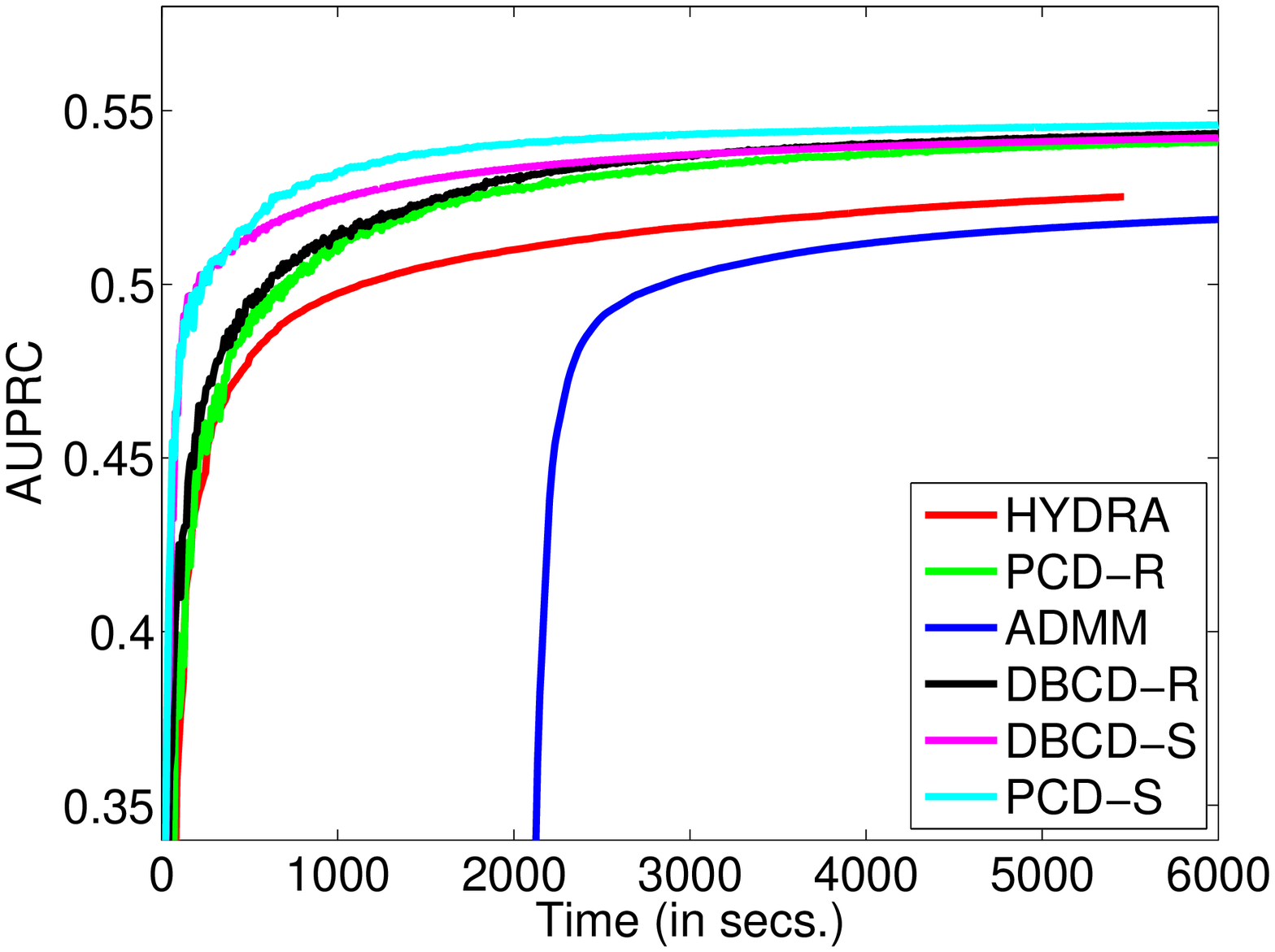}
}
\subfigure[\textsc{URL}, $WSS = 323$]{
\includegraphics[width=0.45\linewidth]{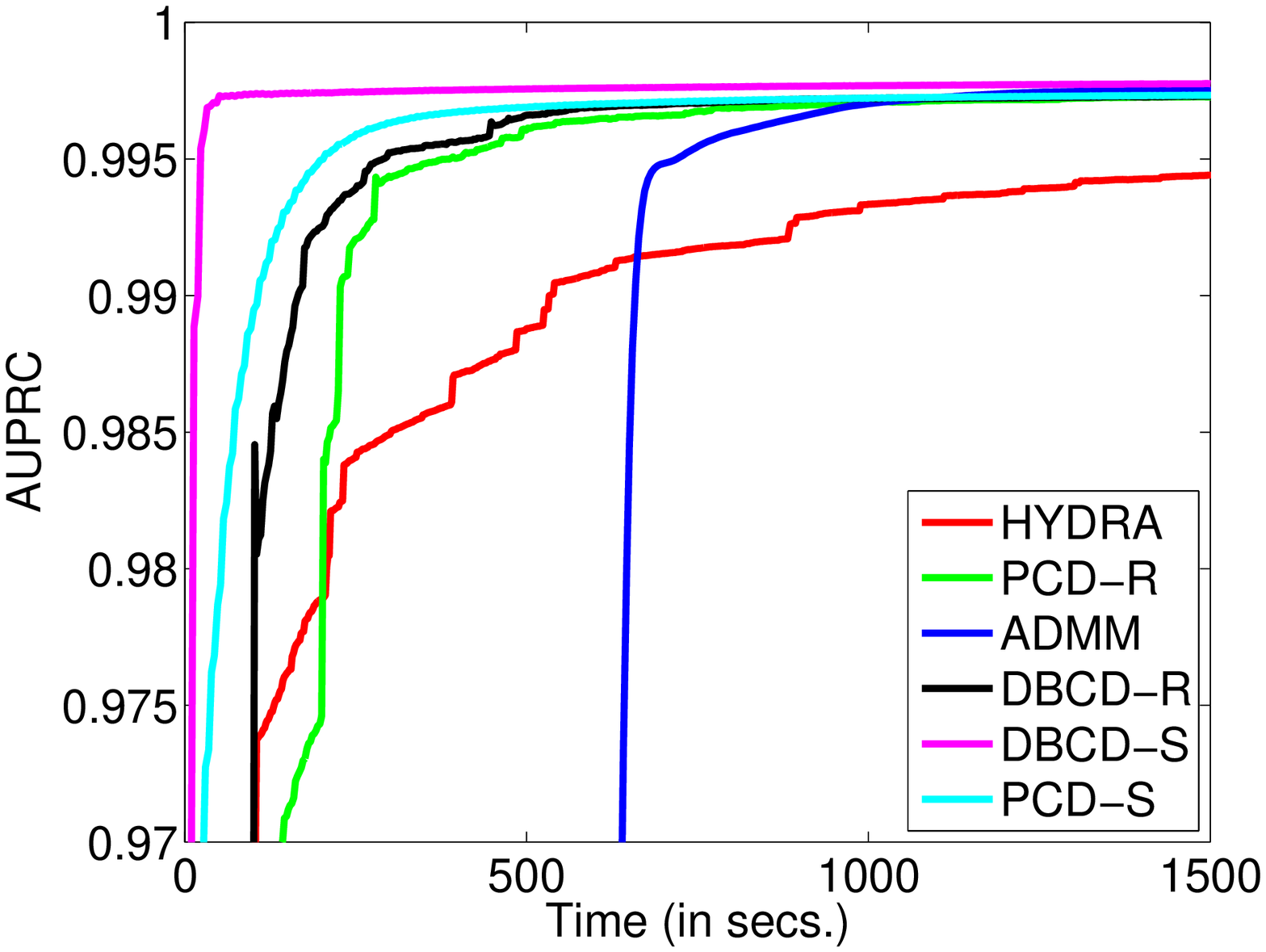}
}
\subfigure[\textsc{URL}, $WSS = 3231$]{
\includegraphics[width=0.45\linewidth]{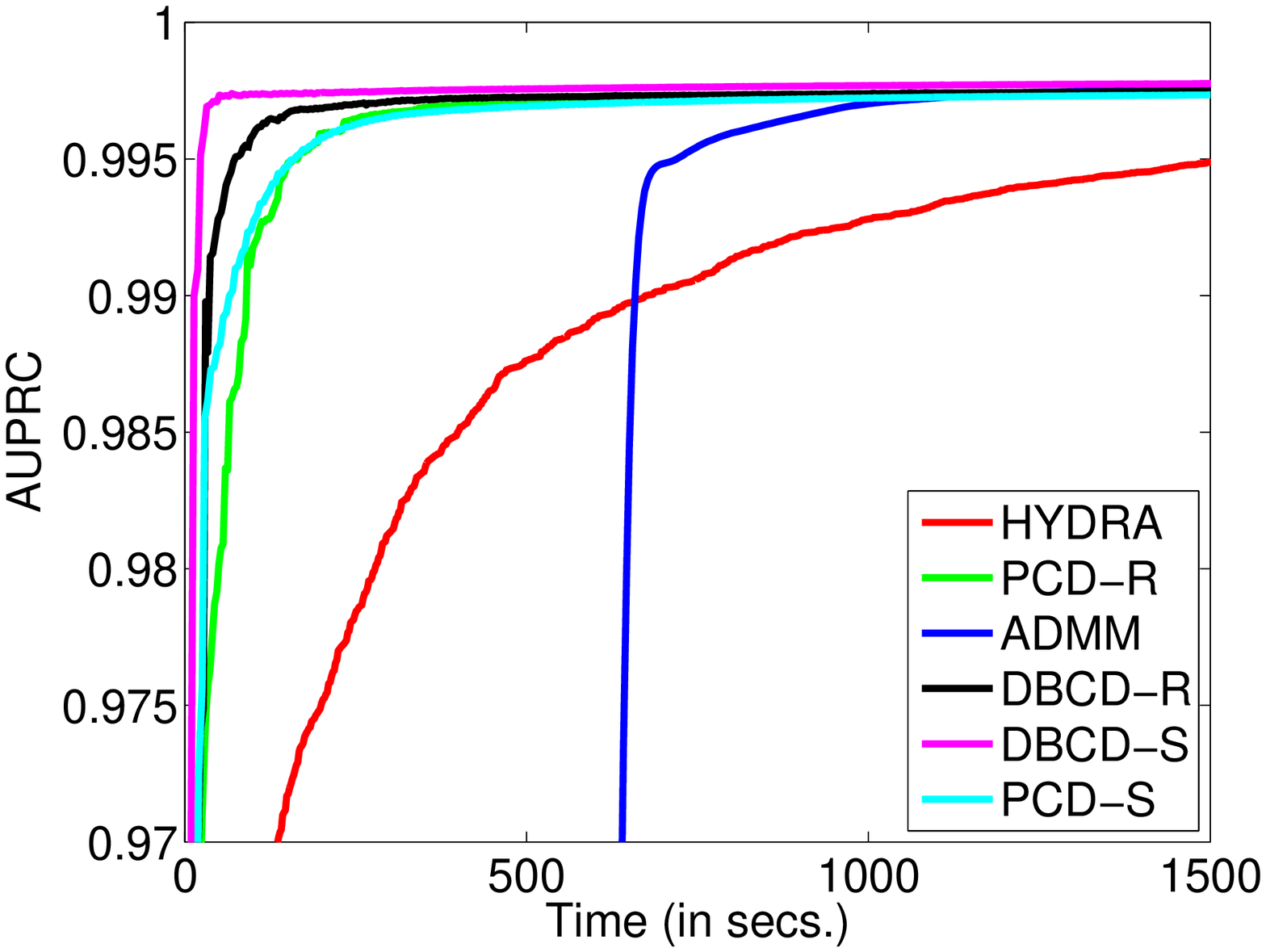}
}
\caption{\small{AUPRC plots. \textsc{KDD} dataset: $\lambda = 1.2 \times 10^{-5}$. \textsc{URL} dataset: $\lambda = 7.3 \times 10^{-7}$}}
\label{fig:kddurlauprc}
\end{figure*}

Figure~\ref{fig:urlobj} shows the objective function plots for (\textit{URL}) with $\lambda$ set to $9 \times 10^{-8}$. Here again, \dbcds~gives the best RFVD performance with order of magnitude speed-up over existing methods. \richtarik~suffers slow convergence and \admm~gives a decent second best performance. Interestingly, the new variable selection rule (S-scheme) did not do very well for \pcdn~for large WSS. This shows that working with quadratic approximations (like \pcdns~does) can at times be quite inferior compared to using the actual nonlinear objective (like \dbcds~does). On comparing the performance for two different WSS, some speed improvement is achieved as in the case of \textsc{KDD} with similar observations. All these objective function progress behaviors are consistent with the AUPRC plots (Figures~\ref{fig:kddauprc} and \ref{fig:urlauprc}) as well except in one case. For example, the AUPRC performance of \pcdns~is quite good although it is a bit slow on the objective function.

Figures~\ref{fig:kddurlobj} and \ref{fig:kddurlauprc} show the performance plots for another choice of different $\lambda$ values for the datasets. \dbcds~gives the best performance on \textit{URL}. On \textit{KDD}, it is the second best after \pcdns. This happens because the $S$ scheme selects features having a large number of non-zero feature values. As a result, computation cost goes up a bit as we do more inner iterations compared to \pcdns. Nevertheless, the performance of \dbcds~is still very good. Overall, the results point to the choice of \dbcds~as the preferred method as it is highly effective with an order of magnitude improvement over existing methods. It is also worth noting that our proposal of using the S-scheme with the \pcdn~method~\citep{bian2013} offers significant value.

Next we analyze the reason behind the superior performance of \dbcds. It is very much along the motivational ideas laid out in Section~\ref{sec:motiv}: since communication cost dominates computation cost in each outer iteration, \dbcds~reduces overall time by decreasing the number of outer iterations. 

\def\stap{{ {> 800} }}

\begin{table}[h]
\caption{$T^P$, the number of outer iterations needed to reach RFVD$\le \tau$, for various $P$ and $\tau$ values. Best values are indicated in boldface. (Note the $\log$ in the definition of RFVD.) Working set size is set at 1\% ($r=0.1$, $WSS=rm/P$). For each dataset, the $\tau$ values were chosen to cover the region where AUPRC values are in the process of reaching the steady state value.}
\label{tab:numiter}
\begin{center}
\begin{tabular}{|c|c|c|c|c|c|c|c|}
\multicolumn{8}{c}{} \\
\multicolumn{8}{c}{KDD, $\lambda = 4.6\times 10^{-7}$} \\
\multicolumn{8}{c}{} \\
\hline
\multicolumn{2}{|c}{} & \multicolumn{3}{|c|}{Existing methods} & \multicolumn{3}{c|}{Our methods}   \\
\hline
P    &  $\tau$     &  \richtarik  &  \admm     &   \pcdnr   &   \pcdns     &  \dbcdr    & \dbcds    \\
\hline
     & $-1$        &  298         & 159        & 294        & 12           & 236        & {\bf 8}   \\
 25  & $-2$        & $\stap$      & 317        & $\stap$    & 311          & $\stap$    & {\bf 104} \\
     & $-3$        & $\stap$      & $\stap$    & $\stap$    & $\stap$      & $\stap$    & {\bf 509} \\
\hline
     & $-1$        & 297          & 180        & 299        & 12           & 230        & {\bf 10}  \\
 100 & $-2$        & $\stap$      & $\stap$    & $\stap$    & 311          & $\stap$    & {\bf 137} \\
     & $-3$        & $\stap$      & $\stap$    & $\stap$    & {\bf 650}    & $\stap$    & 668       \\
\hline
\multicolumn{8}{c}{} \\
\multicolumn{8}{c}{URL, $\lambda = 9.0\times 10^{-8}$} \\
\multicolumn{8}{c}{} \\
\hline
\multicolumn{2}{|c}{} & \multicolumn{3}{|c|}{Existing methods} & \multicolumn{3}{c|}{Our methods}   \\
\hline
     & $0$         & 376          & 137        & 106        & 65           & 101        & {\bf 6}   \\
 25  & $-0.5$      & $\stap$      & 179        & 337        & 117          & 193        & {\bf 14}  \\
     & $-1$        & $\stap$      & 214        & 796        & 196          & 722        & {\bf 22}  \\
\hline
     & $0$         & 400          & 120        & 91         & 64           & 78         & {\bf 7}   \\
 100 & $-0.5$      & $\stap$      & 176        & 313        & 116          & 182        & {\bf 16}  \\
     & $-1$        & $\stap$      & 231        & 718        & 190          & 582        & {\bf 28}  \\
\hline
\end{tabular}
\end{center}
\end{table}

\noindent{\bf Study on the number of outer iterations:} 
We study $T^P$, the number of outer iterations needed to reach RFVD$\le \tau$. Table~\ref{tab:numiter} gives $T^P$ values for various methods in various settings. \dbcds~clearly outperforms other methods in terms of having much smaller values for $T^P$. \pcdns~is the second best method, followed by \admm. The solid reduction of $T^P$ by \dbcds~validates the design that was motivated in Section~\ref{sec:motiv}. The increased computation associated with \dbcds~is immaterial; because communication cost overshadows computation cost in each iteration for all methods, \dbcds~is also the best in terms of the overall computing time. The next set of results gives the details. \vspace*{0.05in}

\noindent{\bf Computation and Communication Time:}
As emphasized earlier, communication plays an important role in the distributed setting. To study this effect, we measured the computation and communication time separately at each node. Figure~\ref{fig:comptime} shows the computation time per node on the \textit{KDD} dataset. In both cases, \admm~incurs significant computation time compared to other methods. This is because it optimizes over all variables in each node. \dbcds~and \dbcdr~come next because our method involves both line search and $10$ inner iterations. \pcdnr~and \pcdns~take a little more time than \richtarik~because of the line search. As seen in both \dbcd~and \pcdn~cases, a marginal increase in time is incurred due to the variable selection cost with the S-scheme compared to the R-scheme.

\begin{table}[h]
\caption{Computation and communication costs per iteration (in secs.) for KDD, $P=25$.}
\label{tab:timings}
\begin{center}
\begin{tabular}{|c|c|c|c|c|c|c|c|c|c}
\hline
Method & Comp. & Comm. & Comp. & Comm.\\
\hline
& \multicolumn{2}{|c|}{WSS: $r=0.1$} & \multicolumn{2}{|c|}{WSS: $r=0.01$}\\
\hline
\richtarik & 0.022 & 5.192 & 0.131 & 4.888\\
\hline
\pcdnr & 0.138 & 5.752 & 0.432 & 5.817\\
\hline
\pcdns & 1.564 & 7.065 & 1.836 & 7.032\\
\hline
\dbcdr & 0.991 & 6.322 & 1.978 & 6.407\\
\hline
\dbcds & 5.054 & 6.563 &  5.557 & 8.867\\
\hline
\end{tabular}
\end{center}
\end{table}

We measured the computation and communication time taken per iteration by each method for different $P$ and $WSS$ settings. From Table~\ref{tab:timings} (which gives representative results for one situation, \textsc{KDD} and $P=25$), we see that the communication time dominates the cost in \richtarik~and \pcdnr.
\dbcdr~takes more computation time than \pcdnr~and \richtarik~since we run through $10$ cycles of inner optimization. Note that the methods with S-scheme take more time; however, the increase is not significant compared to the communication cost. \dbcds~takes the maximum computation time and is quite comparable to the communication time. Recall our earlier observation of \dbcds~giving order of magnitude speed-up in the overall time compared to methods such as \richtarik~and \pcdnr~(see Figures 3-8). Though the computation times taken by \richtarik, \pcdnr~and \pcdns~are lesser, they need significantly more number of iterations to reach some specified objective function optimality criterion. As a result, these methods become quite inefficient due to extremely large communication cost compared to \textsc{DBCD}. All these observations point to the fact our \dbcd~method nicely trades-off the computation versus communication cost, and gives an excellent order of magnitude improvement in overall time. With the additional benefit provided by the S-scheme, \dbcds~clearly turns out to be the method of choice for the distributed setting. \vspace*{0.05in}

\noindent{\bf Sparsity Pattern:} To study weight sparsity behaviors of various methods during optimization, we computed the percentage of non-zero weights ($\rho$) as a function of outer iterations. We set the initial weight vector to zero. Figure~\ref{fig:sparse} shows similar behaviors for all the random (variable) selection methods. After a few iterations of rise they fall exponentially and remain at the same level. For methods with the S-scheme, many variables remain non-zero for some initial period of time and then $\rho$ falls a lot more sharply. It is interesting to note that such an initial behavior seems necessary to make good progress in terms of both function value and AUPRC (Figure~\ref{fig:kddurlobj}(a)(b) and Figure~\ref{fig:kddurlauprc}(a)(b)) In all the cases, many variables stay at zero after initial iterations; therefore, shrinking ideas can be used to improve efficiency. \vspace*{0.05in}

\noindent{\bf Remark on Speed up:}
Let us consider the RFVD plots corresponding to \dbcds~in Figures~\ref{fig:kddobj}~and~\ref{fig:urlobj}. It can be observed that the times associated with $P=25$ and $P=100$ for reaching a certain tolerance, say RFVD=-2, are close to each other. This means that using 100 nodes gives almost no speed up over 25 nodes, which may prompt the question: {\it Is a distributed solution really necessary?} There are two answers to this question. First, as we already mentioned, when the training data is huge\footnote{The \textsc{KDD} and \textsc{URL} datasets are really not huge in the {\it Big data} sense. In this paper we used them only because of lack of availability of much bigger public datasets.} and so {\it the data is generated and forced to reside in distributed nodes}, the right question to ask is not whether we get great speed up, but to ask which method is the fastest. Second, for a given dataset, if the time taken to reach a certain optimality tolerance is plotted as a function of $P$, it may have a minimum at a value different from $P=1$. In such a case, it is appropriate to choose a $P$ (as well as $WSS$) optimally to minimize training time. Many applications involve periodically repeated model training. For example, in Advertising, logistic regression based click probability models are retrained on a daily basis on incrementally varying datasets. In such scenarios it is worthwhile to spend time to tune parameters such as $P$ and $WSS$ in an early deployment phase to minimize time, and then use these parameter values for future runs.

It is also important to point out that the above discussion is relevant to distributed settings in which communication causes a bottleneck. If communication cost is not heavy, e.g., when the number of examples is not large and/or communication is inexpensive such as in multicore solution, then good speed ups are possible; see, for example, the results in~\citet{richtarik2013}.

\begin{figure}[t]
\subfigure[$WSS = 2021$]{
\includegraphics[width=0.48\linewidth]{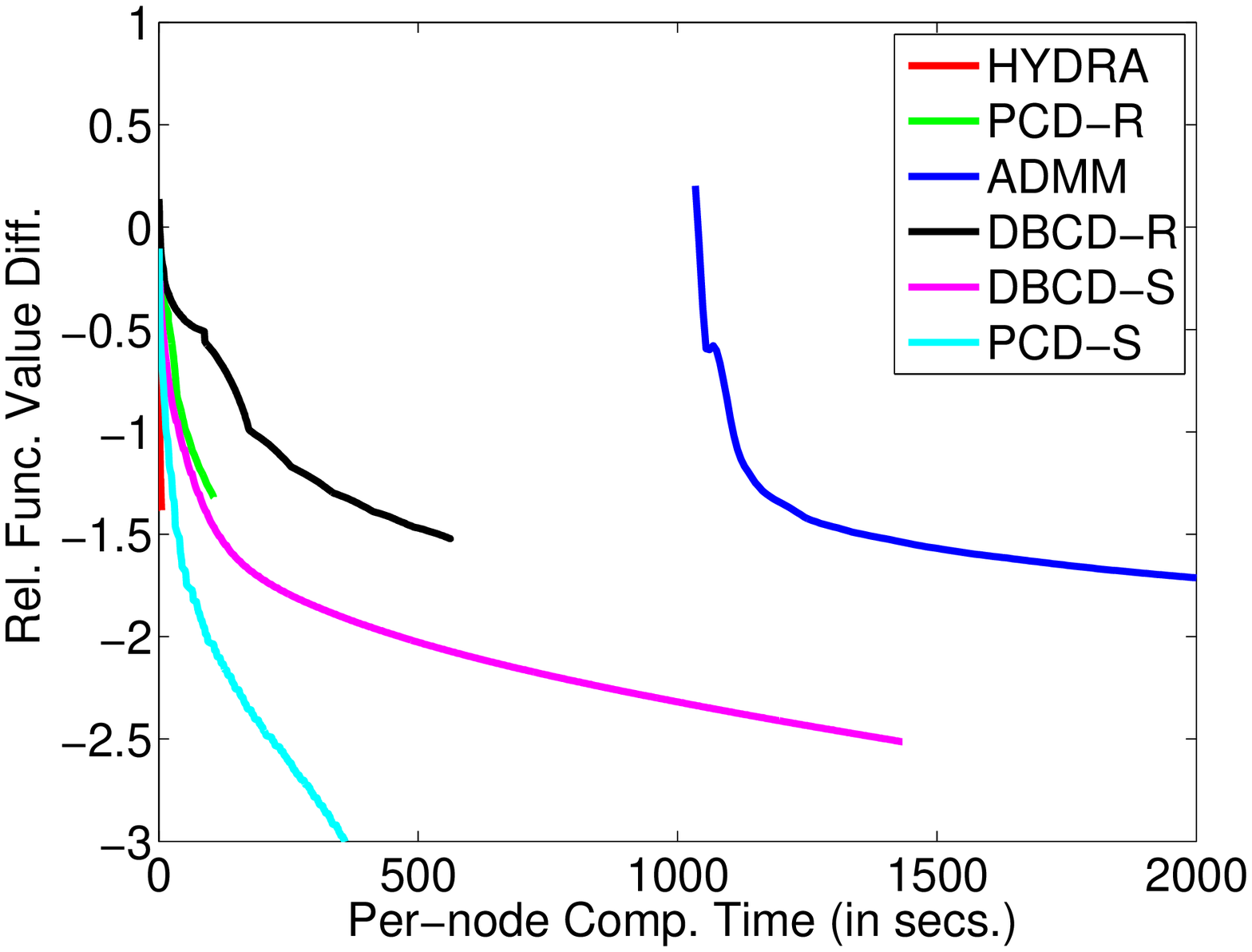}
}
\subfigure[$WSS = 20216$]{
\includegraphics[width=0.48\linewidth]{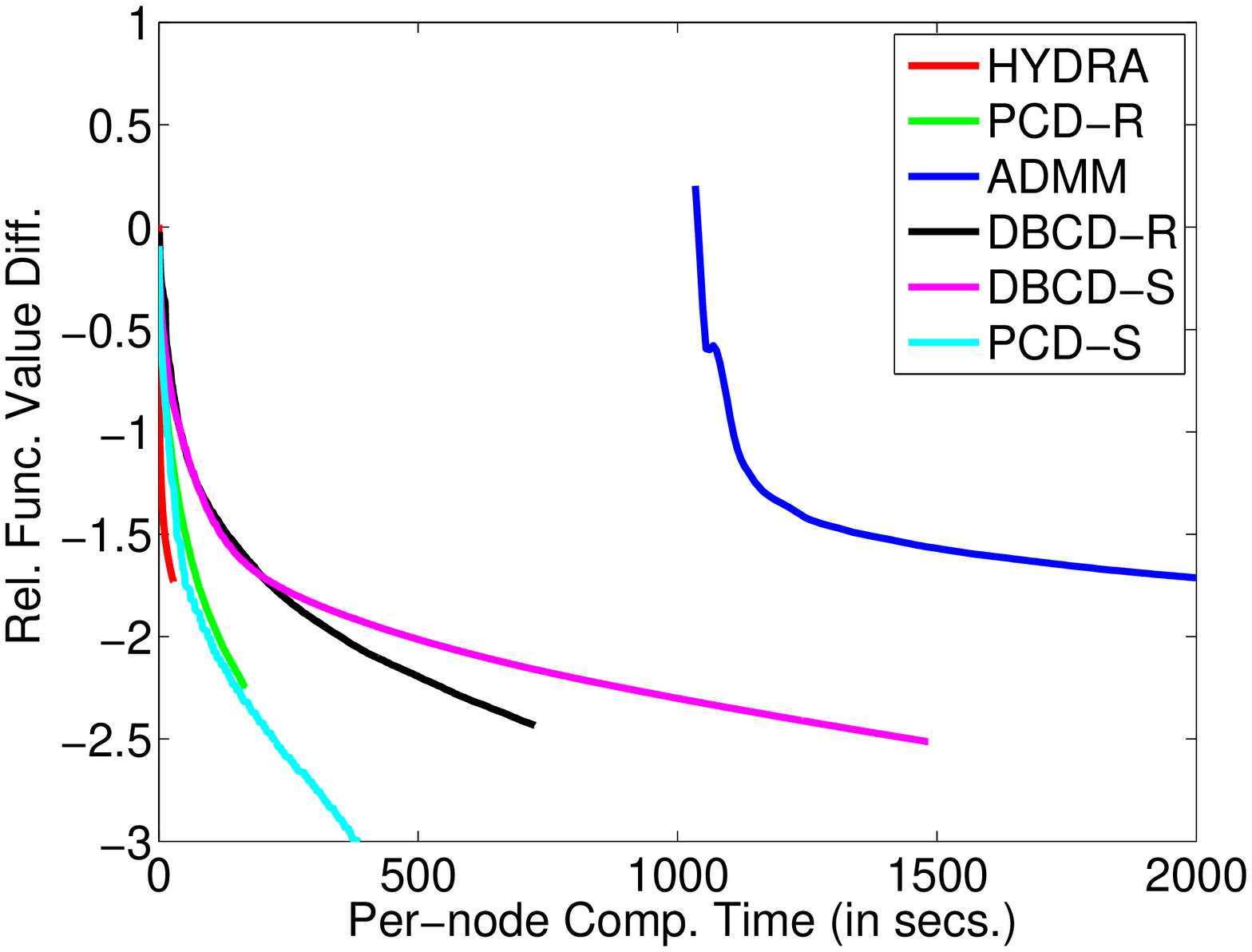}
}
\caption{\small{Per-node computation time on the \textsc{KDD} dataset ($\lambda = 1.2 \times 10^{-5}$ and $P = 100$).}}
\label{fig:comptime}
\end{figure}

\begin{figure}[t]
\subfigure[$WSS = 2021$]{
\includegraphics[width=0.48\linewidth]{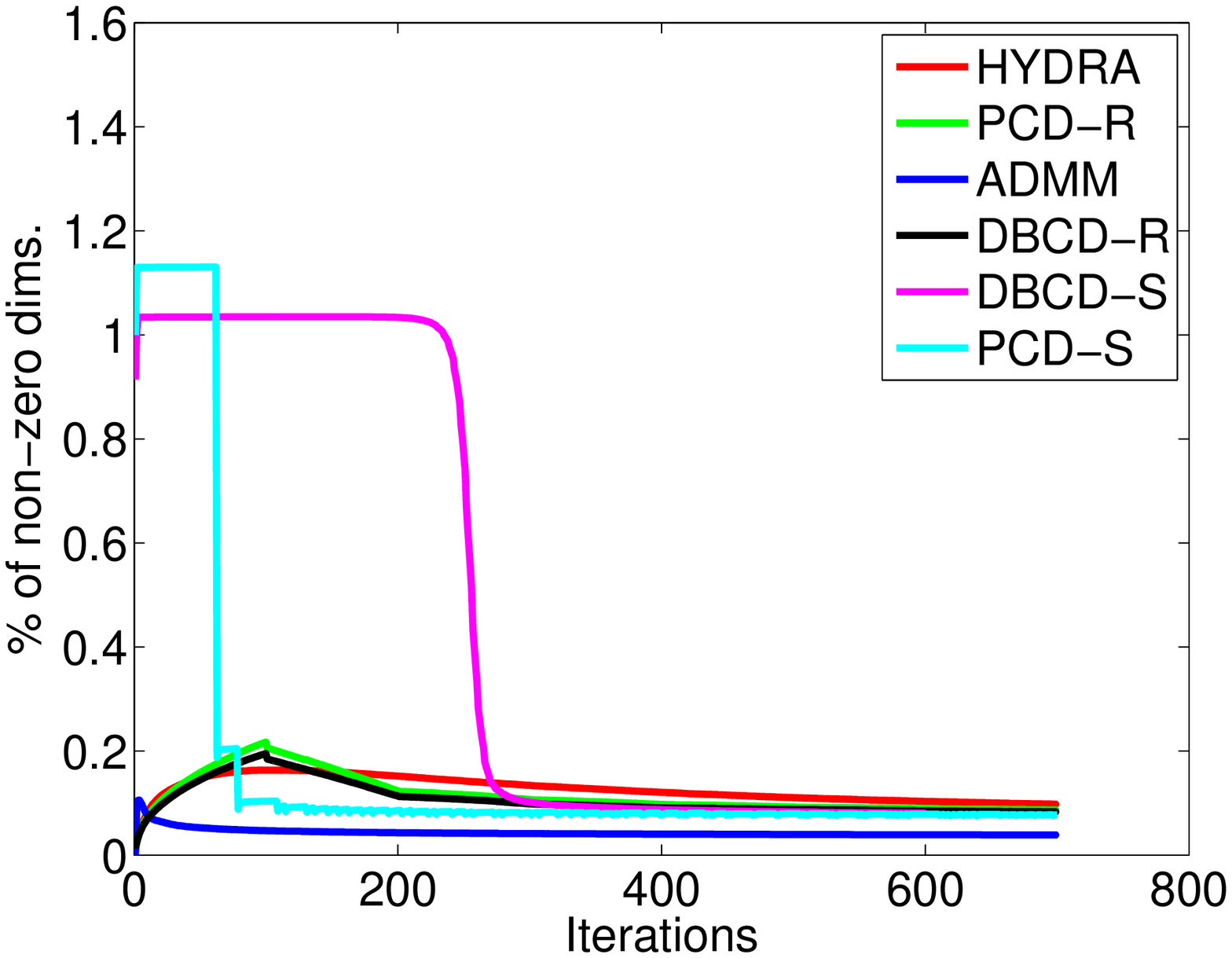}
}
\subfigure[$WSS = 20216$]{
\includegraphics[width=0.48\linewidth]{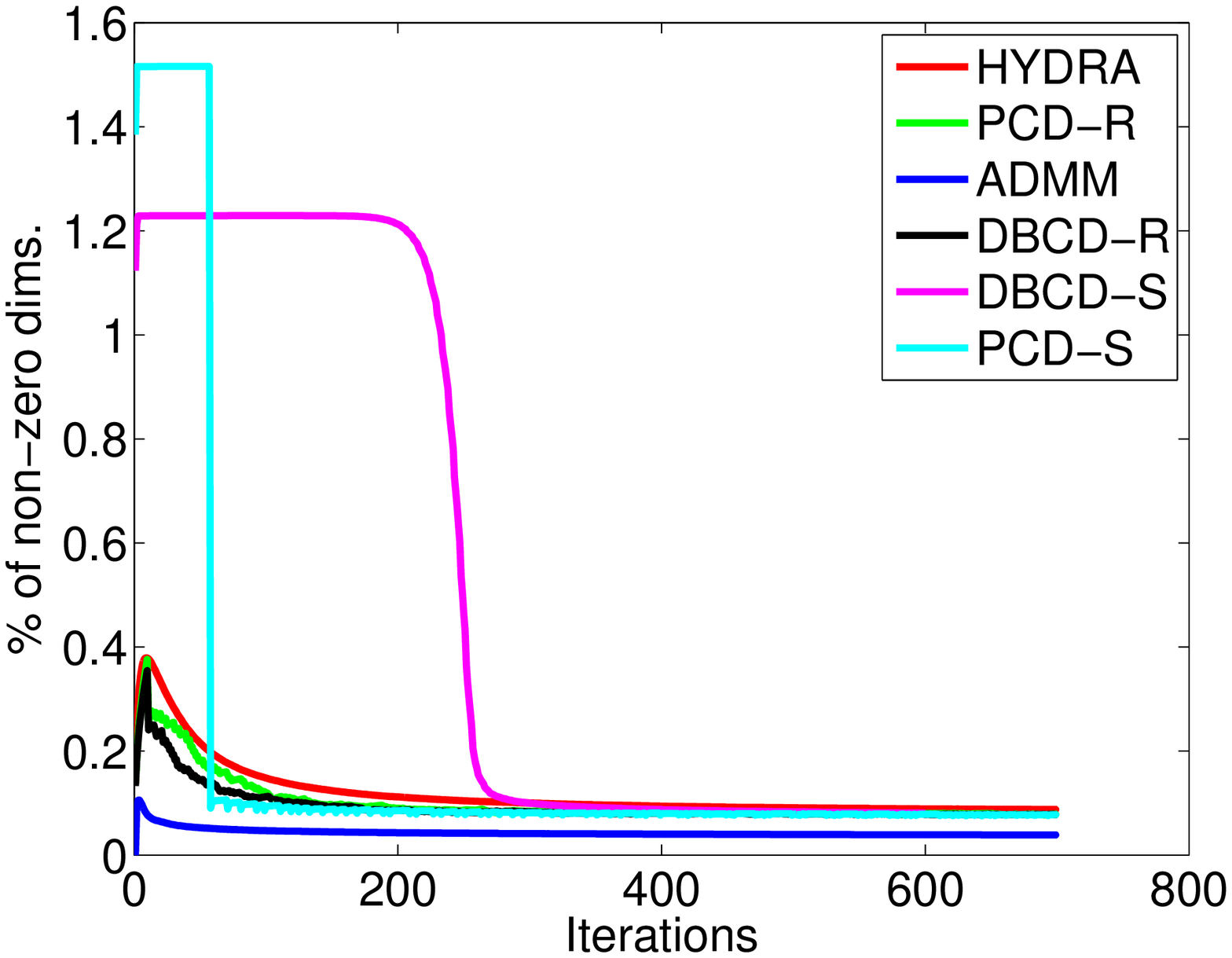}
}
\caption{\small{\textsc{KDD} dataset: Percentage of non-zero weights. $\lambda = 1.2 \times 10^{-5}$ and $P = 100$. }}
\label{fig:sparse}
\end{figure}

\section{Recommended DBCD algorithm}
\label{sec:recom}

In Section~\ref{sec:dbcd} we explored various options for the steps of Algorithm 1 looking beyond those considered by existing methods and proposing new ones, and empirically analyzing the various resulting methods in Section~\ref{sec:expts}. The experiments clearly show that \dbcds~is the best method. We collect full implementation details of this method in~Algorithm~\ref{GAfinal}. 

\begin{algorithm2e}
\caption{Recommended DBCD algorithm\label{GAfinal}}
Parameters:
Proximal constant $\mu>0$ (Default: $\mu=10^{-12}$)\; 
WSS = \# variables to choose for updating per node (Default: WSS=$r\, m/P$, $r=0.1$)\; 
$k=$ \# CD iterations to use for solving~(\ref{Fapprox}) (Default: $k=10$)\;
Line search constants: $\beta,\sigma\in (0,1)$ (Default: $\beta=0.5$, $\sigma=0.01$)\;
Choose $w^0$ and compute $y^0=Xw^0$\;
\For{$t=0,1 \ldots$}{
    \For{$p=1,\ldots, P$ (in parallel)}{
          (a) For each $j\in B_p$, solve~(\ref{quad}) to get $q_j$. Sort $\{ q_j: j\in B_p\}$ and choose WSS indices with least $q_j$ values to form $S_p^t$\;
          (b) Form $f_p^t(\wbp)$ using~(\ref{ours}) and solve~(\ref{Fapprox}) using $k$ CD iterations to get $\wbarpt$ and set direction: $\dbp=\wbarpt-\wbpt$\;
          (c) Compute $\delta y^t = \sum_p \Xbp\dbp$ using AllReduce\;
          (d) $\alpha=1$\;
          \While{(\ref{ls1}-\ref{ls2}) are not satisfied}{
            $\alpha \leftarrow \alpha\beta$\;
            Check (\ref{ls1})-(\ref{ls2}) using $y + \alpha\,\delta y$ and aggregating the $l_1$ regularization value via AllReduce\;
          }
          (e) Set $\alpha^t=\alpha$, $\wbptp = \wbpt + \alpha^t\dbp$ and $y^{t+1} = y^t + \alpha^t \, \delta y^t$\;
     }
     (f) Terminate if the optimality conditions~(\ref{viol}) hold to the desired approximate level\;
}
\end{algorithm2e}

\section{Conclusion}
\label{sec:conc}

In this paper we have proposed a class of efficient block coordinate methods for the distributed training of $l_1$ regularized linear classifiers. In particular, the proximal-Jacobi approximation together with a distributed greedy scheme for variable selection came out as a strong performer. There are several useful directions for the future. It would be useful to explore other approximations such as block GLMNET and block \textsc{L-BFGS} suggested in Subsection 5.1. Like~\citet{richtarik2013}, developing a complexity theory for our method that sheds insight on the effect of various parameters (e.g., $P$) on the number of iterations to reach a specified optimality tolerance is worthwhile. It is possible to extend our method to non-convex problems, e.g., deep net training, which has great value.

%
\bibliography{jmlrfp}  
%
%

	\def\gspt{g_{S_p^t}^t}

\appendix

\section*{Proof of Theorem 1}
\label{app:proof}

First let us write $\delta_j$ in (\ref{appopt}) as $\delta_j = E_{jj} d^t_j$ where $E_{jj}=\delta_j/(\dbp)^j$. Note that $|E_{jj}|\le \mu/2$.
Use the condition (\ref{mvt}) in {\bf P2} with $\wbp=\wbarpt$ and $\wbphat=\wbp$ in (\ref{appopt}) together with the gradient consistency property of {\bf P1} to get
\begin{equation}
\gspt + H_{S_p^t}^t \dsp + \xi_{S_p^t} = 0,
\label{modopt}
\end{equation}
where $H_{S_p^t}^t=\Hhat_{S_p^t}-E_{S_p^t}$ and $\Hhat_{S_p^t}$ is the diagonal submatrix of $\Hhat$ corresponding to $S_p^t$.
Since $\Hhat\ge\mu I$ and $|E_{jj}|\le \mu/2$, we get $H_{S_p^t}^t\ge \frac{\mu}{2}I$.
Let us extend the diagonal matrix $E_{S_p^t}^t$ to $E_{B_p}$ by defining $E_{jj}=0\;\forall j\in B_p\setminus S_p^t$. This lets us extend $H_{S_p^t}^t$ to $H_{B_p}$  via $H_{B_p}^t=\Hhat_{B_p}-E_{B_p}$.

Now (\ref{modopt}) is the optimality condition for the quadratic minimization,
\begin{eqnarray}
\dbp = \arg\min_{d_{B_p}} \;\; (\gbp)^T d_{B_p} + \frac{1}{2} (d_{B_p})^T H_{B_p} d_{B_p}  + \nonumber \\
\sum_{j\in B_p} \lambda\; |w^t_j + d_j |
\;\; \mbox{s.t.} \;\; d_j=0 \; \forall \; j\not\in B_p\setminus{S_p^t}
\label{tsform}
\end{eqnarray}
Combined over all $p$,
\begin{eqnarray}
d^t = \arg\min_d \;\; (g^t)^T d + \frac{1}{2} d^T H d  + u(w^t+d) \nonumber \\
\mbox{s.t.} \;\; d_j=0 \; \forall \; j\not\in \cup_p (B_p\setminus{S_p^t})
\label{tsform2}
\end{eqnarray}
where $H$ is a block diagonal matrix with blocks, $\{H_{B_p}\}$. Thus $d^t$ corresponds to the minimization of a positive definite quadratic form, exactly the type covered by the Tseng-Yun theory~\citep{tseng2009}.

The line search condition (\ref{ls1})-(\ref{ls2}) is a special case of the line search condition in~\citet{tseng2009}.
The Gauss-Seidel scheme of Subsection 5.2 is an instance of the Gauss-Seidel scheme of~\citet{tseng2009}. Now consider the distributed greedy scheme in Subsection 5.2. Let $j_{\max}=\arg\max_{1\le j\le m}{\bar q}_j$. By the way the $S_p^t$ are chosen, $j_{\max}\in \cup_p S_p^t$. Therefore, $\sum_{j\in \cup_p S_p^t} {\bar q}_j \le \frac{1}{m} \sum_{j=1}^m {\bar q}_j$, thus satisfying the Gauss-Southwell-$q$ rule condition of~\citet{tseng2009}.
Now Theorems 1-3 of~\citet{tseng2009} can be directly applied to prove our Theorem 1.

\end{document}